\pdfoutput=1

\documentclass{article}

\usepackage[preprint, nonatbib]{neurips_2022}
\usepackage[utf8]{inputenc} 
\usepackage[T1]{fontenc}    
\usepackage{hyperref}       
\usepackage{url}            
\usepackage{booktabs}       
\usepackage{amsfonts}       
\usepackage{nicefrac}       
\usepackage{microtype}      
\usepackage[table]{xcolor}

\usepackage{xcolor}         
\usepackage{graphicx}
\usepackage{multirow}
\usepackage{array}
\usepackage{amssymb}
\usepackage{amsmath}
\usepackage{caption}
\usepackage{bbding}
\usepackage{subcaption}
\usepackage{blindtext}
\usepackage{adjustbox}
\usepackage{bbm}
\usepackage{sidecap}
\usepackage{threeparttable}
\usepackage{wrapfig}
\input{insbox}

\usepackage{pythonhighlight}

\usepackage{amsmath}

\usepackage{graphicx}
\usepackage[linesnumbered,ruled]{algorithm2e}
\usepackage{algorithmic}

\pdfoutput=1

\usepackage{amsmath,amsfonts,bm}

\def\eqref#1{equation~\ref{#1}}

\def\1{\bm{1}}

\newcommand{\test}{\mathcal{D_{\mathrm{test}}}}

\DeclareMathAlphabet{\mathsfit}{\encodingdefault}{\sfdefault}{m}{sl}
\SetMathAlphabet{\mathsfit}{bold}{\encodingdefault}{\sfdefault}{bx}{n}

\usepackage{amssymb}
\usepackage{pifont}
\newcommand{\cmark}{\ding{51}\xspace}%
\newcommand{\xmarkg}{\textcolor{lightgray}{\ding{55}}\xspace}%
\definecolor{iblue}{rgb}{0.06, 0.75, 1.0}

\newcolumntype{L}[1]{>{\raggedright\let\newline\\\arraybackslash\hspace{0pt}}m{#1}}
\newcolumntype{C}[1]{>{\centering\let\newline\\\arraybackslash\hspace{0pt}}m{#1}}
\newcolumntype{R}[1]{>{\raggedleft\let\newline\\\arraybackslash\hspace{0pt}}m{#1}}
\usepackage[utf8]{inputenc}
\usepackage[english]{babel}
\usepackage{amsthm}
\usepackage[misc]{ifsym}

\newcommand{\app}{\raise.17ex\hbox{$\scriptstyle\sim$}}
\makeatletter\renewcommand\paragraph{\@startsection{paragraph}{4}{\z@}
  {.495em \@plus1ex \@minus.2ex}{-.5em}{\normalfont\normalsize\bfseries}}\makeatother

\makeatletter
\DeclareRobustCommand\onedot{\futurelet\@let@token\@onedot}
\def\@onedot{\ifx\@let@token.\else.\null\fi\xspace}

\def\eg{\emph{e.g}\onedot} 
\def\ie{\emph{i.e}\onedot}

\def\etal{\emph{et al}\onedot}
\newcommand{\one}[1]{\mathbbm{1}_{[#1]}}

\makeatother

\newcolumntype{x}[1]{>{\centering\arraybackslash}p{#1pt}}
\newcolumntype{a}[1]{>{\columncolor{verylightgray}\centering\arraybackslash}p{#1pt}}
\newcolumntype{y}[1]{>{\raggedright\arraybackslash}p{#1pt}}
\newcolumntype{z}[1]{>{\raggedleft\arraybackslash}p{#1pt}}\newlength\savewidth\newcommand\shline{\noalign{\global\savewidth\arrayrulewidth\global\arrayrulewidth 1pt}\hline\noalign{\global\arrayrulewidth\savewidth}}
\newcommand{\tablestyle}[2]{\setlength{\tabcolsep}{#1}\renewcommand{\arraystretch}{#2}\centering\footnotesize}

\newcolumntype{P}[1]{>{\centering\arraybackslash}p{#1}}

\title{Making Your First Choice: To Address\\Cold Start Problem in Vision Active Learning}

\author{
Liangyu Chen\textsuperscript{1} \quad Yutong Bai\textsuperscript{2} \quad Siyu Huang\textsuperscript{3} \quad Yongyi Lu\textsuperscript{2} \quad \\
\bf
Bihan Wen\textsuperscript{1} \quad Alan L. Yuille\textsuperscript{2} \quad Zongwei Zhou\textsuperscript{2,}\thanks{Corresponding author: Zongwei Zhou (\href{mailto:zzhou82@jh.edu}{zzhou82@jh.edu})} \\
\textsuperscript{1}Nanyang Technological University\quad
\textsuperscript{2}Johns Hopkins University\quad
\textsuperscript{3}Harvard University
}

\begin{document}

\maketitle

\setcounter{footnote}{0}

\everypar{\looseness=-1}
\begin{abstract}

    Active learning promises to improve annotation efficiency by iteratively selecting the most important data to be annotated first.
    However, we uncover a striking contradiction to this promise: active learning fails to select data as efficiently as random selection at the first few choices.
    We identify this as the cold start problem in vision active learning, caused by a biased and outlier initial query.
    This paper seeks to address the cold start problem by exploiting the three advantages of contrastive learning:
    (1) no annotation is required; (2) label diversity is ensured by pseudo-labels to mitigate bias; (3) typical data is determined by contrastive features to reduce outliers.
    Experiments are conducted on CIFAR-10-LT and three medical imaging datasets (\ie Colon Pathology, Abdominal CT, and Blood Cell Microscope).
    Our initial query not only significantly outperforms existing active querying strategies but also surpasses random selection by a large margin.
    We foresee our solution to the cold start problem as a simple yet strong baseline to choose the initial query for vision active learning. 
    
    Code is available: \href{https://github.com/c-liangyu/CSVAL}{https://github.com/c-liangyu/CSVAL}


\end{abstract}

\section{Introduction}
\label{sec:introduction}

\begin{center}
``\emph{The secret of getting ahead is getting started.}''
\end{center}
\begin{flushright}
---~Mark Twain
\end{flushright}

\begin{figure}[!t]
\centering
    \includegraphics[width=0.81\linewidth]{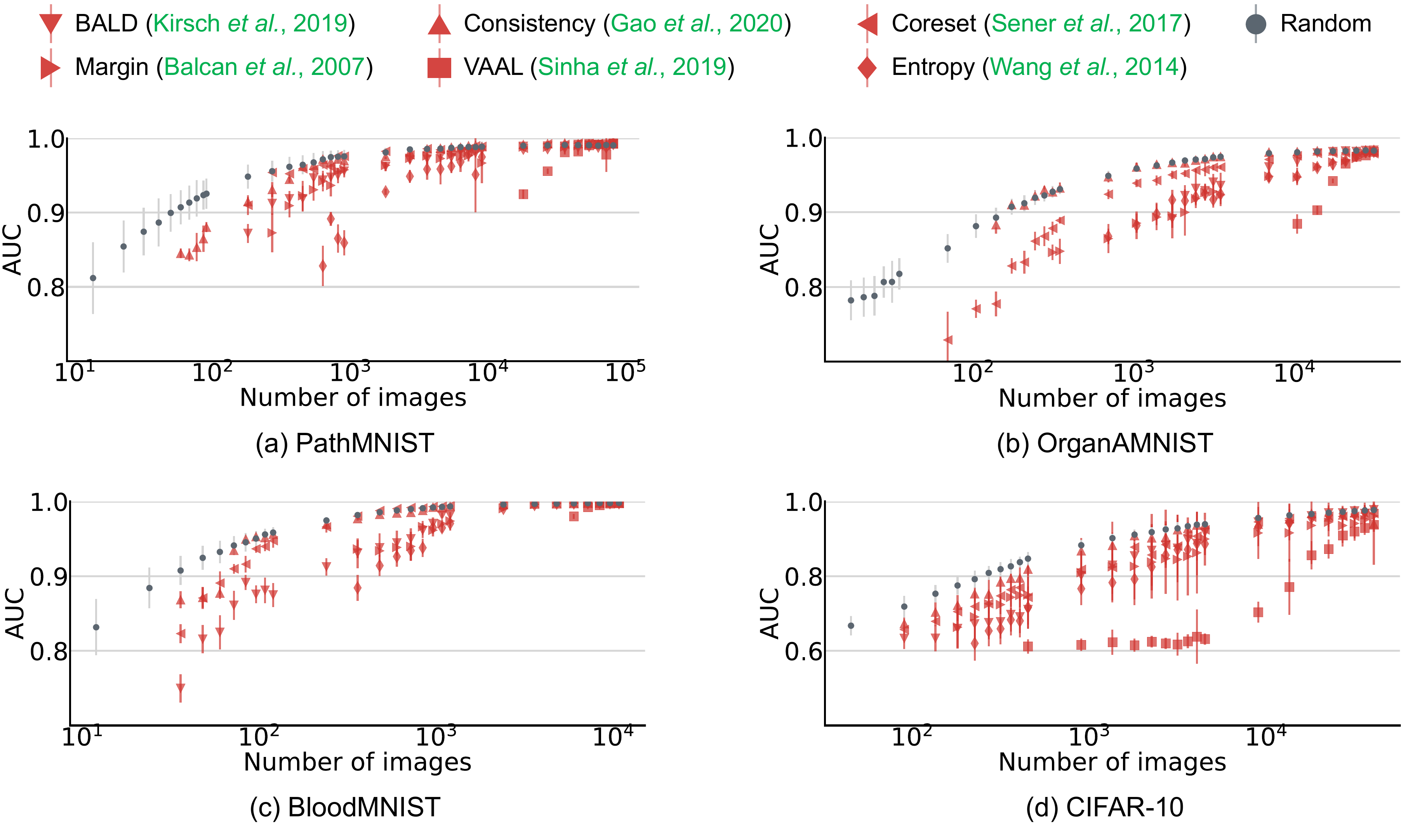}
    \caption{
    \textbf{Cold start problem in vision active learning.} 
    Most existing active querying strategies (\eg BALD, Consistency, etc.) are outperformed by random selection in selecting initial queries, since random selection is i.i.d. to the entire dataset.
    However, some classes are not selected by active querying strategies due to selection bias, so their results are not presented in the low budget regime.
    }
    \label{fig:teaser_cold_start}
\end{figure}

The cold start problem was initially found in recommender systems~\cite{zhang2010solving,qiu2011item,bobadilla2012collaborative,houlsby2014cold} when algorithms had not gathered sufficient information about users with no purchase history.
It also occurred in many other fields, such as natural language processing~\cite{yuan2020cold,margatina2021bayesian} and computer vision~\cite{bengar2021reducing,chandra2021initial,pourahmadi2021simple} during the active learning procedure\footnote{Active learning aims to select the most important data from the unlabeled dataset and query human experts to annotate new data. The newly annotated data is then added to improve the model. This process can be repeated until the model reaches a satisfactory performance level or the annotation budget is exhausted.}. 
Active learning promises to improve annotation efficiency by iteratively selecting the most important data to annotate.
However, we uncover a striking contradiction to this promise: Active learning fails to select data as effectively as random selection at the first choice.
We identify this as the cold start problem in vision active learning and illustrate the problem using three medical imaging applications (\figureautorefname~\ref{fig:teaser_cold_start}a--c) as well as a natural imaging application (\figureautorefname~\ref{fig:teaser_cold_start}d).
Cold start is a crucial topic~\cite{yehuda2022active,lang2021best} because a performant initial query can lead to noticeably improved subsequent cycle performance in the active learning procedure, evidenced in~\S\ref{sec:better_better}.
There is a lack of studies that systematically illustrate the cold start problem, investigate its causes, and provide practical solutions to address it. To this end, we ask: {\em What causes the cold start problem and how can we select the initial query when there is no labeled data available?}

Random selection is generally considered a baseline to start the active learning because the randomly sampled query is independent and identically distributed (i.i.d.) to the entire data distribution.
As is known, maintaining a similar distribution between training and test data is beneficial, particularly when using limited training data~\cite{jadon2021covid}.
Therefore, a large body of existing work selects the initial query randomly~\cite{borisov2010active,zhou2017fine,yuan2020cold,zhou2021active,gao2020consistency,gal2017deep,settles2009active,huang2021semi,holub2008entropy,zhou2019integrating}, highlighting that active querying compromises accuracy and diversity compared to random sampling at the beginning of active learning ~\cite{mittal2019parting,zhu2019addressing,simeoni2021rethinking,chandra2021initial,hacohen2022active,zhou2021towards}.
Why?
We attribute the causes of the cold start problem to the following two aspects:

(i) \textit{Biased query}: Active learning tends to select data that is biased to specific classes.
Empirically, \figureautorefname~\ref{fig:selected_label_histogram_pathmnist} reveals that the class distribution in the selected query is highly unbalanced. These active querying strategies (\eg Entropy, Margin, VAAL, etc.) can barely outperform random sampling at the beginning because some classes are simply not selected for training.
It is because data of the minority classes occurs much less frequently than those of the majority classes.
Moreover, datasets in practice are often highly unbalanced, particularly in medical images~\cite{litjens2017survey,zhou2021review}.
This can escalate the biased sampling.
We hypothesize that the \textit{label diversity} of a query is an important criterion to determine the importance of the annotation. 
To evaluate this hypothesis theoretically, we explore the upper bound performance by enforcing a uniform distribution using ground truth (\tableautorefname~\ref{tab:label_diversity})
To evaluate this hypothesis practically, we pursue the label diversity by exploiting the pseudo-labels generated by \textit{K}-means clustering (\tableautorefname~\ref{tab:label_coverage}).
The label diversity can reduce the redundancy in the selection of majority classes, and increase the diversity by including data of minority classes.

(ii) \textit{Outlier query}: Many active querying strategies were proposed to select typical data and eliminate outliers, but they heavily rely on a trained classifier to produce predictions or features. For example, to calculate the value of Entropy, a trained classifier is required to predict logits of the data. However, there is no such classifier at the start of active learning, at which point no labeled data is available for training. 
To express informative features for reliable predictions, we consider contrastive learning, which can be trained using unlabeled data only. Contrastive learning encourages models to discriminate between data augmented from the same image and data from different images~\cite{chen2020improved,chen2020simple}. Such a learning process is called instance discrimination.
We hypothesize that instance discrimination can act as an alternative to select typical data and eliminate outliers.
Specifically, the data that is hard to discriminate from others could be considered as typical data.
With the help of Dataset Maps~\cite{swayamdipta2020dataset,karamcheti2021mind}\footnote{It is worthy noting that both~\cite{swayamdipta2020dataset} and~\cite{karamcheti2021mind} conducted a retrospective study, which analyzed existing active querying strategies by using the ground truth. As a result, the values of \textit{confidence} and \textit{variability} in the Dataset Maps could not be computed under the practical active learning setting because the ground truth is a priori unknown. Our modified strategy, however, does not require the availability of ground truth (detailed in~\S\ref{sec:method_hard_to_contrast}).}, we evaluate this hypothesis and propose a novel active querying strategy that can effectively select \textit{typical data} (\textit{hard-to-contrast} data in our definition, see~\S\ref{sec:method_hard_to_contrast}) and reduce outliers.

Systematic ablation experiments and qualitative visualizations in \S\ref{sec:experiment} confirm that (i) the level of label diversity and (ii) the inclusion of typical data are two explicit criteria for determining the annotation importance.
Naturally, contrastive learning is expected to approximate these two criteria: pseudo-labels in clustering implicitly enforce label diversity in the query;
instance discrimination determines typical data.
Extensive results show that our initial query not only significantly outperforms existing active querying strategies, but also surpasses random selection by a large margin on three medical imaging datasets (\ie Colon Pathology, Abdominal CT, and Blood Cell Microscope) and two natural imaging datasets (\ie CIFAR-10 and CIFAR-10-LT).
Our active querying strategy eliminates the need for manual annotation to ensure the label diversity within initial queries, and more importantly, starts the active learning procedure with the typical data. 

To the best of our knowledge, we are among the first to indicate and address the cold start problem in the field of medical image analysis (and perhaps, computer vision), making three contributions:
(1) illustrating the cold start problem in vision active learning,
(2) investigating the underlying causes with rigorous empirical analysis and visualization, and
(3) determining effective initial queries for the active learning procedure.
Our solution to the cold start problem can be used as a strong yet simple baseline to select the initial query for image classification and other vision tasks.



\noindent\textbf{Related work.}
When the cold start problem was first observed in recommender systems, there were several solutions to remedy the insufficient information due to the lack of user history~\cite{zhu2019addressing,houlsby2014cold}.
In natural language processing (NLP), Yuan~\etal~\cite{yuan2020cold} were among the first to address the cold start problem by pre-training models using self-supervision.
They attributed the cold start problem to model instability and data scarcity.
Vision active learning has shown higher performance than random selection~\cite{zhou2017fine,sourati2019intelligent,gao2020consistency,Agarwal2020ContextualDF,shui2020deep,mayer2020adversarial,zhou2021active}, but there is limited study discussing how to select the initial query when facing the entire unlabeled dataset.
A few studies somewhat indicated the existence of the cold start problem:
Lang~\etal~\cite{lang2021best} explored the effectiveness of the $K$-center algorithm~\cite{farahani2009facility} to select the initial queries.
Similarly, Pourahmadi~\etal~\cite{pourahmadi2021simple} showed that a simple $K$-means clustering algorithm worked fairly well at the beginning of active learning, as it was capable of covering diverse classes and selecting a similar number of data per class.
Most recently, a series of studies~\cite{hacohen2022active,yehuda2022active,sorscher2022beyond,nath2022warm} continued to propose new strategies for selecting the initial query from the entire unlabeled data and highlighted that typical data (defined in varying ways) could significantly improve the learning efficiency of active learning at a low budget.
In addition to the existing publications, our study justifies the two causes of the cold start problem, systematically presents the existence of the problem in six dominant strategies, and produces a comprehensive guideline of initial query selection.

\section{Method}
\label{sec:method}

In this section, we analyze in-depth the cause of cold start problem in two perspectives, biased query as the inter-class query and outlier query as the intra-class factor. We provide a complementary method to select the initial query based on both criteria.
\S\ref{sec:method_label_diversity} illustrates that label diversity is a favourable selection criterion, and discusses how we obtain label diversity via simple contrastive learning and $K$-means algorithms. 
\S\ref{sec:method_hard_to_contrast} describes an unsupervised method to sample atypical (hard-to-contrast) queries from Dataset Maps.

\subsection{Inter-class Criterion: Enforcing Label Diversity to Mitigate Bias}
\label{sec:method_label_diversity}

\begin{figure}[!t]
    \includegraphics[width=0.9\linewidth]{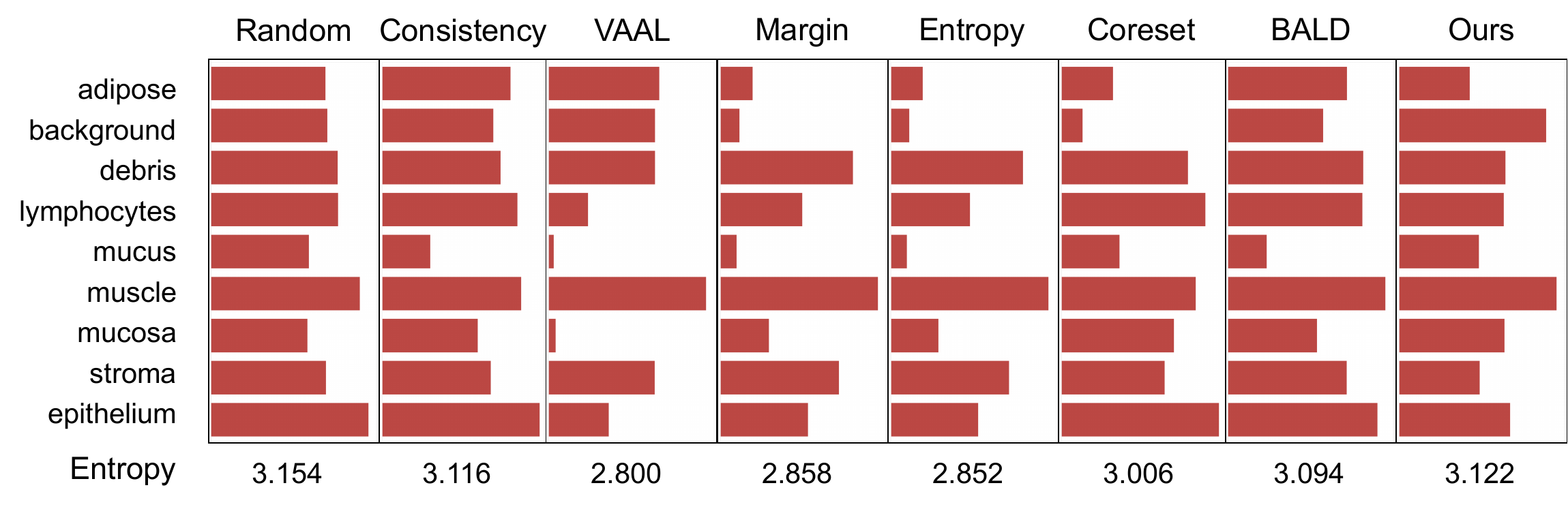}
    \caption{
        \textbf{Label diversity of querying criteria.}
        Random, the leftmost strategy, denotes the class distribution of randomly queried samples, which can also reflect the approximate class distribution of the entire dataset.
        As seen, even with a relatively larger initial query budget (40,498 images, 45\% of the dataset), most active querying strategies are biased towards certain classes in the PathMNIST dataset.
        For example, VAAL prefers selecting data in the muscle class, but largely ignores data in the mucus and mucosa classes.
        On the contrary, our querying strategy selects more data from minority classes (e.g., mucus and mucosa) while retaining the class distribution of major classes.
        Similar observations in OrganAMNIST and BloodMNIST are shown in Appendix~\figureautorefname~\ref{fig:selected_label_histogram_appendix}. The higher the entropy is, the more balanced the class distribution is.
    }
    \label{fig:selected_label_histogram_pathmnist}
    \vspace{-3mm}
\end{figure}


\noindent\textbf{$K$-means clustering.}
The selected query should cover data of diverse classes, and ideally, select similar number of data from each class. 
However, this requires the availability of ground truth, which are inaccessible according to the nature of active learning.
Therefore, we exploit pseudo-labels generated by a simple $K$-means clustering algorithm and select an equal number of data from each cluster to form the initial query to facilitate label diversity.
Without knowledge about the exact number of ground-truth classes, over-clustering is suggested in recent works~\cite{van2020scan,zheltonozhskii2020self} to increase performances on the datasets with higher intra-class variance. Concretely, given 9, 11, 8 classes in the ground truth, we set $K$ (the number of clusters) to 30 in our experiments.

\noindent\textbf{Contrastive features.}
$K$-means clustering requires features of each data point. 
Li~\etal~\cite{li2021contrastive} suggested that for the purpose of clustering, contrastive methods (\eg MoCo, SimCLR, BYOL) are more suitable than generative methods (\eg colorization, reconstruction) because the contrastive feature matrix can be naturally regarded as cluster representations.
Therefore, we use MoCo v2~\cite{chen2020improved}---a popular self-supervised contrastive method---to extract image features. 

$K$-means and MoCo~v2 are certainly not the only choices for clustering and feature extraction.
We employ these two well-received methods for simplicity and efficacy in addressing the cold start problem. 
\figureautorefname~\ref{fig:selected_label_histogram_pathmnist} shows our querying strategy can yield better label diversity than other six dominant active querying strategies;
similar observations are made in OrganAMNIST and BloodMNIST (\figureautorefname~\ref{fig:selected_label_histogram_appendix}) as well as CIFAR-10 and CIFAR-10-LT (\figureautorefname~\ref{fig:selected_label_histogram_cifar_appendix}).

\subsection{Intra-class Criterion: Querying Hard-to-Contrast Data to Avoid Outliers}
\label{sec:method_hard_to_contrast}

\begin{figure}[!t]
    \includegraphics[width=1.0\linewidth]{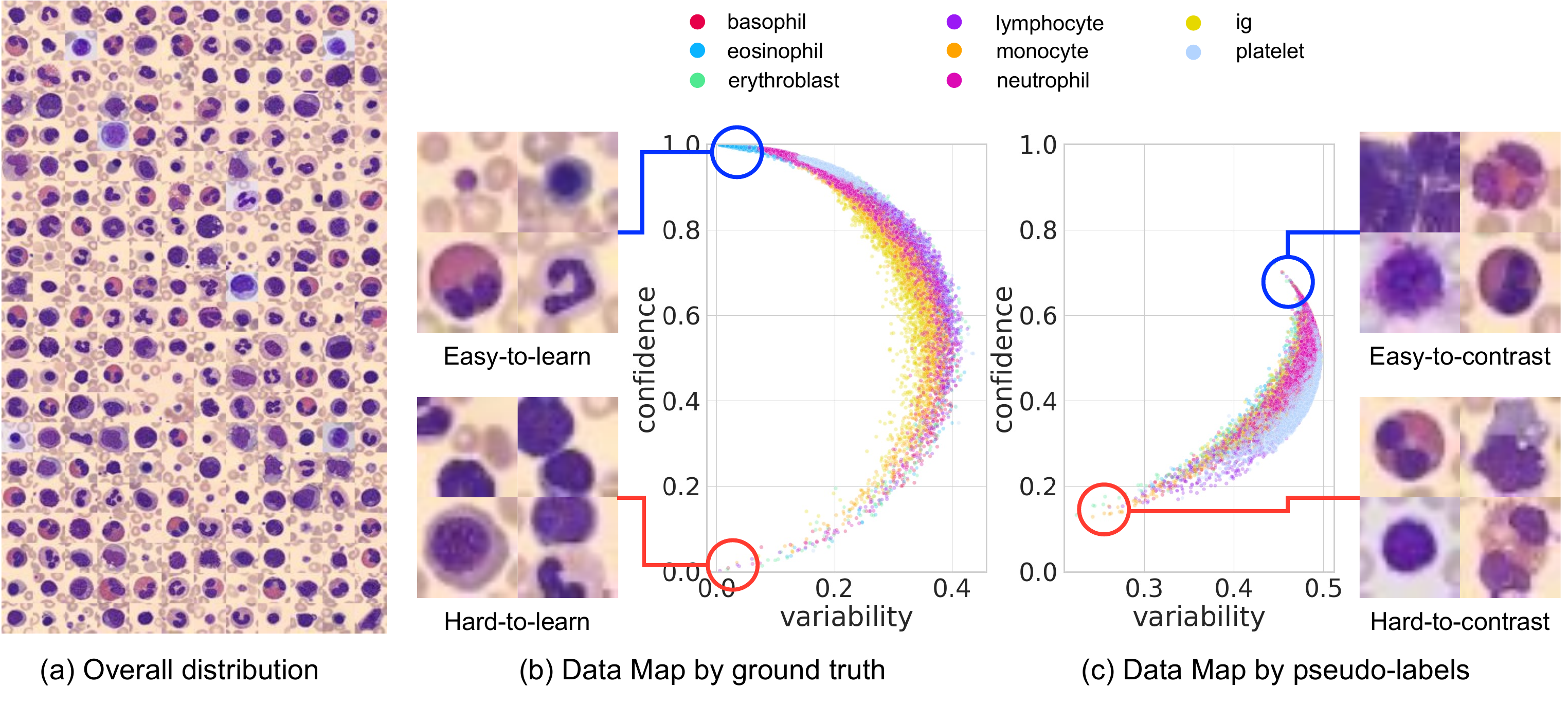}
    \caption{\textbf{Active querying based on Dataset Maps.}
        (a) Dataset overview.
        (b) Easy- and hard-to-learn data can be selected from the maps based on ground truths~\cite{karamcheti2021mind}. This querying strategy has two limitations: it requires manual annotations and the data are stratified by classes in the 2D space, leading to a poor label diversity in the selected queries.
        (c) Easy- and hard-to-contrast data can be selected from the maps based on pseudo-labels.
        This querying strategy is label-free and the selected hard-to-contrast data represent the most common patterns in the entire dataset, as presented in (a). These data are more suitable for training, and thus alleviate the cold start problem.
    }
    \label{fig:cartography_all}
\end{figure}

\noindent\textbf{Dataset map.}
Given $K$ clusters generated from Criterion~\#1, we now determine which data points ought to be selected from each cluster.
Intuitively, a data point can better represent a cluster distribution if it is harder to contrast itself with other data points in this cluster---we consider them typical data.
To find these typical data, we modify the original Dataset Map\footnote{Dataset Map~\cite{chang2017active,swayamdipta2020dataset} was proposed to analyze datasets by two measures: \textit{confidence} and \textit{variability}, defined as the mean and standard deviation of the model probability of ground truth along the learning trajectory.} by replacing the ground truth term with a pseudo-label term.
This modification is made because ground truths are unknown in the active learning setting but pseudo-labels are readily accessible from Criterion~\#1.
For a visual comparison, \figureautorefname~\ref{fig:cartography_all}b and \figureautorefname~\ref{fig:cartography_all}c present the Data Maps based on ground truths and pseudo-labels, respectively.
Formally, the modified Data Map can be formulated as follows.
Let $\mathcal{D} = \{{{\bm x}_m}\}_{m=1}^{M}$ denote a dataset of $M$ unlabeled images.
Considering a minibatch of $N$ images, for each image ${\bm x}_n$, its two augmented views form a positive pair, denoted as $\tilde{\bm x}_i$ and $\tilde{\bm x}_j$. The contrastive prediction task on pairs of augmented images derived from the minibatch generate $2N$ images, in which a true label $y_n^*$ for an anchor augmentation is associated with its counterpart of the positive pair. We treat the other $2(N-1)$ augmented images within a minibatch as negative pairs. We define the probability of positive pair in the instance discrimination task as:
\begin{equation}
    p_{i,j} = \frac{\exp(\mathrm{sim}(\bm z_i,\bm z_j))/\tau}{\sum_{n=1}^{2N}\one{n \neq i}\exp(\mathrm{sim}(\bm z_i,\bm z_n))/\tau},
    \label{eq:p_ij}
\end{equation}

\begin{equation}
    p_{\theta^{(e)}}(y_n^*|x_n)=\frac{1}{2}[p_{2n-1, 2n}+p_{2n, 2n-1}],
    \label{eq:p_theta_e}
\end{equation}

where $\mathrm{sim}(\bm u,\bm ) = \bm u^\top \bm v / \lVert\bm u\rVert \lVert\bm v\rVert$ is the cosine similarity between $\bm u$ and $\bm v$; $\bm z_{2n-1}$ and $\bm z_{2n}$ denote the projection head output of a positive pair for the input $\bm x_n$ in a batch; $\one{n \neq i} \in \{ 0,  1\}$ is an indicator function evaluating to $1$ iff $n \neq i$ and $\tau$ denotes a temperature parameter. ${\theta}^{(e)}$ denotes the parameters at the end of the $e^\text{th}$ epoch.
We define confidence $(\hat{\mu}_m)$ across $E$ epochs as:
\begin{equation}
    \hat{\mu}_m=\frac{1}{E}\sum_{e=1}^{E}p_{\theta^{(e)}}(y_m^*|x_m).
    \label{eq:confidence}
\end{equation}
The confidence $(\hat{\mu}_m)$ is the Y-axis of the Dataset Maps (see~\figureautorefname~\ref{fig:cartography_all}b-c).

\noindent\textbf{Hard-to-contrast data.}
We consider the data with a low confidence value (\equationautorefname~\ref{eq:confidence}) as ``hard-to-contrast'' because they are seldom predicted correctly in the instance discrimination task.
Apparently, if the model cannot distinguish a data point with others, this data point is expected to carry typical characteristics that are shared across the dataset~\cite{robinson2020contrastive}.
Visually, hard-to-contrast data gather in the bottom region of the Dataset Maps and ``easy-to-contrast'' data gather in the top region. As expected, hard-to-learn data are more typical, possessing the most common visual patterns as the entire dataset; whereas easy-to-learn data appear like outliers~\cite{yehuda2022active,karamcheti2021mind}, which may not follow the majority data distribution (examples in~\figureautorefname~\ref{fig:cartography_all}a and \figureautorefname~\ref{fig:cartography_all}c).
Additionally, we also plot the original Dataset Map~\cite{chang2017active,swayamdipta2020dataset} in~\figureautorefname~\ref{fig:cartography_all}b, which grouped data into hard-to-learn and easy-to-learn\footnote{Swayamdipta~\etal~\cite{swayamdipta2020dataset} indicated that easy-to-learn data facilitated model training in the low budget regime because easier data reduced the confusion when the model approaching the rough decision boundary. In essence, the advantage of easy-to-learn data in active learning aligned with the motivation of curriculum learning~\cite{Bengio2009CurriculumL}.}. Although the results in~\S\ref{sec:experiment_hard_to_contrast} show equally compelling performance achieved by both easy-to-learn~\cite{swayamdipta2020dataset} and hard-to-contrast data (ours), the latter do not require any manual annotation, and therefore are more practical and suitable for vision active learning.

In summary, to meet the both criteria, our proposed active querying strategy includes three steps: (i) extracting features by self-supervised contrastive learning, (ii) assigning clusters by $K$-means algorithm for label diversity, and (iii) selecting hard-to-contrast data from dataset maps.

\section{Experimental Results}
\label{sec:experiment}

\begin{table*}[!t]
    \footnotesize
    \centering
    \caption{
        \textbf{Diversity is a significant add-on to most querying strategies.}
        AUC scores of different querying strategies are compared on three medical imaging datasets.
        In either low budget (\ie 0.5\% or 1\% of MedMNIST datasets) or high budget (\ie 10\% or 20\% of CIFAR-10-LT) regimes, both random and active querying strategies benefit from enforcing the label diversity of the selected data.
        The cells are highlighted in blue when adding diversity performs no worse than the original querying strategies.
        Coreset~\cite{sener2017active} works very well as its original form because this querying strategy has implicitly considered the label diversity (also verified in~\tableautorefname~\ref{tab:label_coverage}) by formulating a $K$-center problem, which selects $K$ data points to represent the entire dataset.
        Some results are missing (marked as ``-'') because the querying strategy fails to sample at least one data point for each class.
        Results of more sampling ratios are presented in Appendix~Figures~\ref{fig:label_diversity_appendix},~\ref{fig:label_diversity_cifar_appendix}.
    }
    \label{tab:label_diversity}
    \scalebox{0.8}{
        \begin{tabular}{p{0.09\linewidth}P{0.06\linewidth}|P{0.1\linewidth}P{0.1\linewidth}|P{0.1\linewidth}P{0.1\linewidth}|P{0.1\linewidth}P{0.1\linewidth} |P{0.1\linewidth}P{0.1\linewidth}@{}}
            \toprule
                                         &         & \multicolumn{2}{c}{PathMNIST}    & \multicolumn{2}{c}{OrganAMNIST}  & \multicolumn{2}{c}{BloodMNIST}   & \multicolumn{2}{c}{CIFAR-10-LT}                                                                                                                                               \\
                                         &         & 0.5\%                            & 1\%                              & 0.5\%                            & 1\%                              & 0.5\%                            & 1\%                               & 10\%                             & 20\%                             \\
                                         & Unif.   & (499)                            & (899)                            & (172)                            & (345)                            & (59)                             & (119)                             & (1420)                           & (2841)                           \\
            \shline
            \multirow{2}{*}{Random}      & \cmark  & \cellcolor{iblue!30}96.8$\pm$0.6 & \cellcolor{iblue!30}97.6$\pm$0.6 & \cellcolor{iblue!30}91.1$\pm$0.9 & \cellcolor{iblue!30}93.3$\pm$0.4 & \cellcolor{iblue!30}94.7$\pm$0.7 & \cellcolor{iblue!30}96.5$\pm$0.4  & \cellcolor{iblue!30}91.6$\pm$1.1 & \cellcolor{iblue!30}93.1$\pm$0.6 \\
                                         & \xmarkg & 96.4$\pm$1.3                     & 97.6$\pm$0.9                     & 90.7$\pm$1.1                     & 93.1$\pm$0.7                     & 93.2$\pm$1.5                     & 95.8$\pm$0.7                      & 62.0$\pm$6.1                     & -                                \\
            \hline
            \multirow{2}{*}{Consistency} & \cmark  & \cellcolor{iblue!30}96.4$\pm$0.1 & \cellcolor{iblue!30}97.9$\pm$0.1 & \cellcolor{iblue!30}92.3$\pm$0.5 & 92.8$\pm$1.0                     & \cellcolor{iblue!30}92.9$\pm$0.9 & \cellcolor{iblue!30}95.9$\pm$0.5  & \cellcolor{iblue!30}91.4$\pm$1.1 & \cellcolor{iblue!30}93.4$\pm$0.2 \\
                                         & \xmarkg & 96.2$\pm$0.0                     & 97.6$\pm$0.0                     & 91.0$\pm$0.3                     & 94.0$\pm$0.6                     & 87.9$\pm$0.2                     & 95.5$\pm$0.5                      & 67.1$\pm$17.1                    & 88.6$\pm$0.3                     \\
            \hline
            \multirow{2}{*}{VAAL}        & \cmark  & \cellcolor{iblue!30}92.7$\pm$0.5 & \cellcolor{iblue!30}93.0$\pm$0.6 & \cellcolor{iblue!30}70.6$\pm$1.9 & \cellcolor{iblue!30}84.6$\pm$0.5 & \cellcolor{iblue!30}89.8$\pm$1.3 & \cellcolor{iblue!30}93.4$\pm$0. 9 & \cellcolor{iblue!30}92.6$\pm$0.2 & \cellcolor{iblue!30}93.7$\pm$0.4 \\
                                         & \xmarkg & -                                & -                                & -                                & -                                & -                                & -                                 & -                                & -                                \\
            \hline
            \multirow{2}{*}{Margin}      & \cmark  & \cellcolor{iblue!30}97.9$\pm$0.2 & \cellcolor{iblue!30}96.0$\pm$0.4 & \cellcolor{iblue!30}81.8$\pm$1.2 & 85.8$\pm$1.4                     & \cellcolor{iblue!30}89.7$\pm$1.9 & \cellcolor{iblue!30}94.7$\pm$0.7  & \cellcolor{iblue!30}91.7$\pm$0.9 & \cellcolor{iblue!30}93.2$\pm$0.2 \\
                                         & \xmarkg & 91.0$\pm$2.3                     & 96.0$\pm$0.3                     & -                                & 85.9$\pm$0.7                     & -                                & -                                 & 81.9$\pm$0.8                     & 86.3$\pm$0.3                     \\
            \hline
            \multirow{2}{*}{Entropy}     & \cmark  & \cellcolor{iblue!30}93.2$\pm$1.6 & \cellcolor{iblue!30}95.2$\pm$0.2 & \cellcolor{iblue!30}79.1$\pm$2.3 & \cellcolor{iblue!30}86.7$\pm$0.8 & \cellcolor{iblue!30}85.9$\pm$0.5 & \cellcolor{iblue!30}91.8$\pm$1.0  & \cellcolor{iblue!30}92.0$\pm$1.2 & \cellcolor{iblue!30}91.9$\pm$1.3 \\
                                         & \xmarkg & -                                & 87.5$\pm$0.1                     & -                                & -                                & -                                & -                                 & 65.6$\pm$15.6                    & 86.4$\pm$0.2                     \\
            \hline
            \multirow{2}{*}{Coreset}     & \cmark  & 95.0$\pm$2.2                     & 94.8$\pm$2.5                     & \cellcolor{iblue!30}85.6$\pm$0.4 & \cellcolor{iblue!30}89.9$\pm$0.5 & \cellcolor{iblue!30}88.5$\pm$0.6 & \cellcolor{iblue!30}94.1$\pm$1.1  & \cellcolor{iblue!30}91.5$\pm$0.4 & \cellcolor{iblue!30}93.6$\pm$0.2 \\
                                         & \xmarkg & 95.6$\pm$0.7                     & 97.5$\pm$0.2                     & 83.8$\pm$0.6                     & 88.5$\pm$0.4                     & 87.3$\pm$1.6                     & 94.0$\pm$1.2                      & 65.9$\pm$15.9                    & 86.9$\pm$0.1                     \\
            \hline
            \multirow{2}{*}{BALD}        & \cmark  & \cellcolor{iblue!30}95.8$\pm$0.2 & \cellcolor{iblue!30}97.0$\pm$0.1 & \cellcolor{iblue!30}87.2$\pm$0.3 & \cellcolor{iblue!30}89.2$\pm$0.3 & \cellcolor{iblue!30}89.9$\pm$0.8 & 92.7$\pm$0.7                      & \cellcolor{iblue!30}92.8$\pm$0.1 & \cellcolor{iblue!30}90.8$\pm$2.4 \\
                                         & \xmarkg & 92.0$\pm$2.3                     & 95.3$\pm$1.0                     & -                                & -                                & 83.3$\pm$2.2                     & 93.5$\pm$1.3                      & 64.9$\pm$14.9                    & 84.7$\pm$0.6                     \\
            \bottomrule
        \end{tabular}
    }
\end{table*}

\begin{table*}[!t]
    \footnotesize
    \centering
    \caption{\textbf{Class coverage of selected data.}
        Compared with random selection (i.i.d. to entire data distribution), most active querying strategies contain selection bias to specific classes, so the class coverage in their selections might be poor, particularly using low budgets.
        As seen, using 0.002\% or even smaller proportion of MedMNIST datasets, the class coverage of active querying strategies is much lower than random selection.
        By integrating $K$-means clustering with contrastive features, our querying strategy is capable of covering 100\% classes in most scenarios using low budgets ($\leq$0.002\% of MedMNIST). We also found that our querying strategy covers the most of the classes in the CIFAR-10-LT dataset, which is designatedly more imbalanced.
    }
    \label{tab:label_coverage}
    \scalebox{0.8}{
        \begin{tabular}{p{0.2\linewidth}|P{0.1\linewidth}P{0.1\linewidth}|P{0.1\linewidth}P{0.1\linewidth}|P{0.1\linewidth}P{0.1\linewidth}|P{0.1\linewidth}P{0.1\linewidth}@{}}
            \toprule
                        & \multicolumn{2}{c}{PathMNIST} & \multicolumn{2}{c}{OrganAMNIST} & \multicolumn{2}{c}{BloodMNIST} & \multicolumn{2}{c}{CIFAR-10-LT}                                                                 \\
                        & 0.00015\%                     & 0.00030\%                       & 0.001\%                        & 0.002\%                         & 0.001\%       & 0.002\%       & 0.2\%         & 0.3\%         \\
                        & (13)                          & (26)                            & (34)                           & (69)                            & (11)          & (23)          & (24)          & (37)          \\
            \shline
            Random      & \textbf{0.79$\pm$0.11}                 & 0.95$\pm$0.07                   & 0.91$\pm$0.08                  & 0.98$\pm$0.04                   & 0.70$\pm$0.13 & 0.94$\pm$0.08 & 0.58$\pm$0.10 & 0.66$\pm$0.12 \\
            \hline
            Consistency & 0.78                          & 0.88                            & 0.82                           & 0.91                            & 0.75          & 0.88          & 0.50          & 0.70          \\
            VAAL        & 0.11                          & 0.11                            & 0.18                           & 0.18                            & 0.13          & 0.13          & 0.30          & 0.30          \\
            Margin      & 0.67                          & 0.78                            & 0.73                           & 0.82                            & 0.63          & 0.75          & 0.60          & 0.70          \\
            Entropy     & 0.33                          & 0.33                            & 0.45                           & 0.73                            & 0.63          & 0.63          & 0.40          & 0.70          \\
            Coreset     & 0.66                          & 0.78                            & 0.91                           & 1.00                            & 0.63          & 0.88          & 0.60          & 0.70          \\
            BALD        & 0.33                          & 0.44                            & 0.64                           & 0.64                            & 0.75          & 0.88          & 0.60          & 0.70          \\
            \hline
            Ours        & 0.78                          & \textbf{1.00}                            & \textbf{1.00}                           & \textbf{1.00}                            & \textbf{1.00}          & \textbf{1.00}          & \textbf{0.70}          & \textbf{0.80}          \\
            \bottomrule
        \end{tabular}
    }
\end{table*}

\noindent\textbf{Datasets \& metrics.} Active querying strategies have a selection bias that is particularly harmful in long-tail distributions. Therefore, unlike most existing works~\cite{pourahmadi2021simple,yehuda2022active}, which tested on highly balanced annotated datasets, we deliberately examine our method and other baselines on long-tail datasets to simulate real-world scenarios. Three medical datasets of different modalities in MedMNIST~\cite{medmnistv2} are used: PathMNIST (colorectal cancer tissue histopathological images), BloodMNIST (microscopic peripheral blood cell images), OrganAMNIST (axial view abdominal CT images of multiple organs).
OrganAMNIST is augmented following Azizi~\etal~\cite{azizi2021big}, while the others following Chen~\etal~\cite{chen2020improved}.
Area Under the ROC Curve (AUC) and Accuracy are used as the evaluation metrics.
All results were based on at least three independent runs, and particularly, 100 independent runs for random selection.
UMAP~\cite{mcinnes2018umap} is used to analyze feature clustering results.

\noindent\textbf{Baselines \& implementations.}
We benchmark a total of seven querying strategies: (1) random selection, (2) Max-Entropy~\cite{wang2014new}, (3) Margin~\cite{balcan2007margin}, (4) Consistency~\cite{gao2020consistency}, (5) BALD~\cite{kirsch2019batchbald}, (6) VAAL~\cite{sinha2019variational}, and (7) Coreset~\cite{sener2017active}.
For contrastive learning, we trained 200 epochs with MoCo~v2, following its default hyperparameter settings. We set $\tau$ to 0.05 in \eqref{eq:p_theta_e}. To reproduce the large batch size and iteration numbers in~\cite{chen2020simple}, we apply repeated augmentation~\cite{Hoffer2020AugmentYB,tong2022videomae,Touvron2021TrainingDI} (detailed in \tableautorefname~\ref{tab:repeat_augmentation}).
More baseline and implementation details can be found in Appendix~\ref{sec:implementation_appendix}.

\subsection{Contrastive Features Enable Label Diversity to Mitigate Bias}

\noindent\textbf{Label coverage \& diversity.}
Most active querying strategies have selection bias towards specific classes, thus the class coverage in their selections might be poor (see~\tableautorefname~\ref{tab:label_coverage}), particularly at low budgets.
By simply enforcing label diversity to these querying strategies can significantly improve the performance (see~\tableautorefname~\ref{tab:label_diversity}), which suggests that the label diversity is one of the causes that existing active querying strategies perform poorer than random selection. 

Our proposed active querying strategy, however, is capable of covering 100\% classes in most low budget scenarios ($\leq$0.002\% of full dataset) by integraing $K$-means clustering with contrastive features.

\subsection{Pseudo-labels Query Hard-to-Contrast Data and Avoid Outliers}
\label{sec:experiment_hard_to_contrast}

\begin{figure}[!t]
    \includegraphics[width=\linewidth]{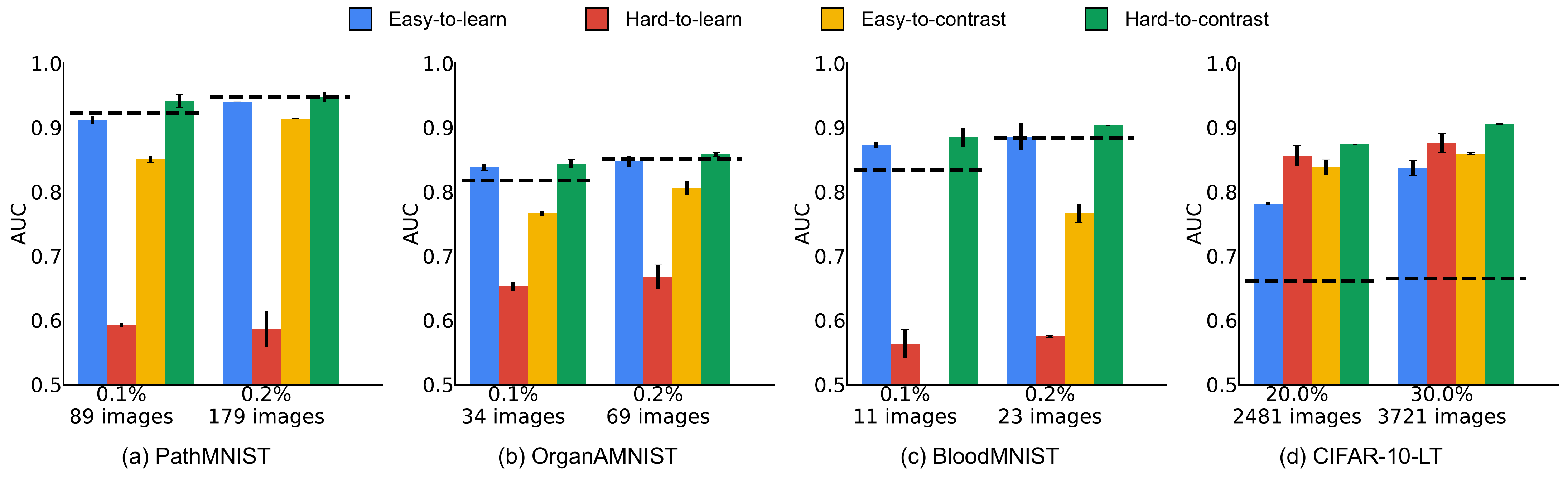}
    \caption{
        \textbf{Quantitative comparison of map-based querying strategies.}
        Random selection (dot-lines) can be treated as a highly competitive baseline in cold start because it outperforms six popular active querying strategies as shown in~\figureautorefname~\ref{fig:teaser_cold_start}.
        In comparison with random selection and three other querying strategies, hard-to-contrast performs the best.
        Although easy-to-learn and hard-to-learn sometimes performs similarly to hard-to-contrast, their selection processes require ground truths~\cite{karamcheti2021mind}, which are  not available in the setting of active learning.
    }
    \label{fig:histogram}
\end{figure}

\noindent\textbf{Hard-to-contrast data are practical for cold start problem.}
\figureautorefname~\ref{fig:histogram} presents the quantitative comparison of four map-based querying strategies, wherein easy- or hard-to-learn are selected by the maps based on ground truths, easy- or hard-to-contrast are selected by the maps based on pseudo-labels.
Note that easy- or hard-to-learn are enforced with label diversity, due to their class-stratified distributions in the projected 2D space (illustrated in~\figureautorefname~\ref{fig:cartography_all}).
Results suggest that \textit{selecting easy-to-learn or hard-to-contrast data contribute to the optimal models}.
In any case, easy- or hard-to-learn data can not be selected without knowing ground truths, so these querying strategies are not practical for active learning procedure.
Selecting hard-to-contrast, on the other hand, is a label-free strategy and yields the highest performance amongst existing active querying strategies (reviewed in~\figureautorefname~\ref{fig:teaser_cold_start}).
More importantly, hard-to-contrast querying strategy significantly outperforms random selection by 1.8\% (94.14\%$\pm$1.0\%~vs.~92.27\%$\pm$2.2\%), 2.6\% (84.35\%$\pm$0.7\%~vs.~81.75\%$\pm$2.1\%), and 5.2\% (88.51\%$\pm$1.5\%~vs.~83.36\%$\pm$3.5\%) on PathMNIST, OrganAMNIST, and BloodMNIST, respectively, by querying 0.1\% of entire dataset. Similarly on CIFAR-10-LT, hard-to-contrast significantly outperforms random selection by 21.2\% (87.35\%$\pm$0.0\%~vs.~66.12\%$\pm$0.9\%) and 24.1\% (90.59\%$\pm$0.1\%~vs.~66.53\%$\pm$0.5\%) by querying 20\% and 30\% of entire dataset respectively.
Note that easy- or hard-to-learn are not enforced with label diversity, for a more informative comparison.


\begin{figure}[!t]
    \includegraphics[width=\linewidth]{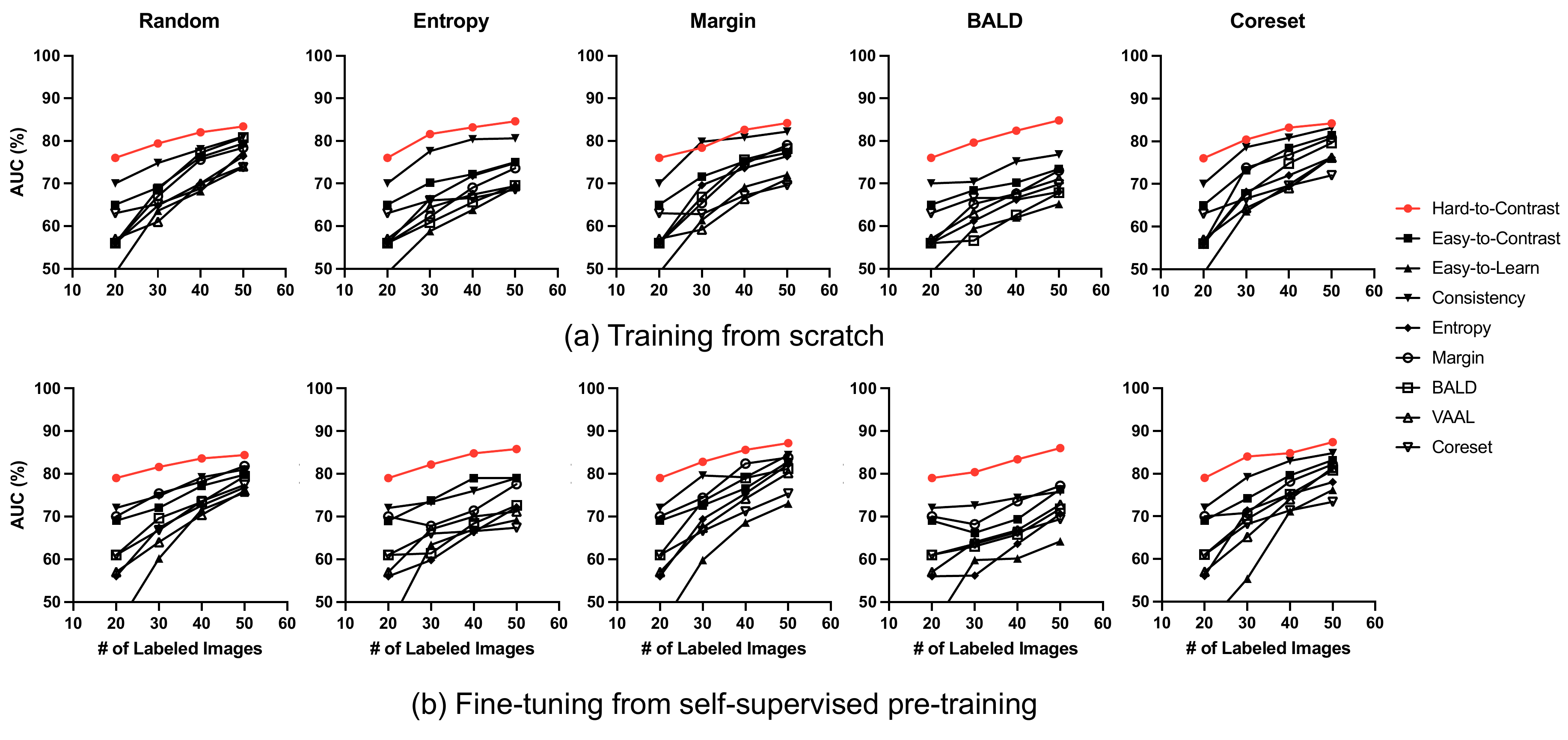}
    \caption{
        \textbf{On the importance of selecting a superior initial query.}
        Hard-to-contrast data (red lines) outperform other initial queries in every cycle of active learning on OrganaMNIST. We find that the performance of the initial cycle (20 images) and the last cycle (50 images) are strongly correlated. 
    }
    \label{fig:better_better_organamnist}
\end{figure}

\subsection{On the Importance of Selecting Superior Initial Query}
\label{sec:better_better}

\noindent\textbf{A good start foresees improved active learning.}
We stress the importance of the cold start problem in vision active learning by conducting correlation analysis. 
Starting with 20 labeled images as the initial query, the training set is increased by 10 more images in each active learning cycle.
\figureautorefname~\ref{fig:better_better_pathmnist}a presents the performance along the active learning (each point in the curve accounts for 5 independent trials).
The initial query is selected by a total of 9 different strategies\footnote{Hard-to-learn is omitted because it falls behind other proposed methods by a large margin (\figureautorefname~\ref{fig:histogram}).}, and subsequent queries are selected by 5 different strategies.
AUC$_n$ denotes the AUC score achieved by the model that is trained by $n$ labeled images. 
The Pearson correlation coefficient between AUC$_{20}$ (starting) and AUC$_{50}$ (ending) shows strong positive correlation ($r=$ 0.79, 0.80, 0.91, 0.67, 0.92 for random selection, Entropy, Margin, BALD, and Coreset, respectively). This result is statistically significant ($p<$ 0.05).
Hard-to-contrast data (our proposal) consistently outperforms the others on OrganAMNIST (\figureautorefname~\ref{fig:better_better_organamnist}), BloodMNIST (\figureautorefname~\ref{fig:better_better_bloodmnist}), and PathMNIST (\figureautorefname~\ref{fig:better_better_pathmnist}), and steadily improves the model performances within the next active learning cycles.




\noindent\textbf{The initial query is consequential regardless of model initialization.}
A pre-trained model can improve the performance of each active learning cycle for both random and active selection~\cite{yuan2020cold},
but the cold start problem remains (evidenced in~\figureautorefname~\ref{fig:better_better_pathmnist}b).
This suggests that the model instability and data scarcity are two independent issues to be addressed for the cold start problem.
Our ``hard-to-contrast'' data selection criterion only exploits contrastive learning (an improved model), but also determines the typical data to be annotated first (a better query).
As a result, when fine-tuning from MoCo~v2, the Pearson correlation coefficient between AUC$_{20}$ and AUC$_{50}$ remains high  ($r=$ 0.92, 0.81, 0.70, 0.82, 0.85 for random selection, Entropy, Margin, BALD, and Coreset, respectively) and statistically significant ($p<$ 0.05).

\section{Conclusion}
\label{sec:conclusion}

This paper systematically examines the causes of the cold start problem in vision active learning and offers a practical and effective solution to address this problem.
Analytical results indicate that (1) the level of label diversity and (2) the inclusion of hard-to-contrast data are two explicit criteria to determine the annotation importance.
To this end, we devise a novel active querying strategy that can enforce label diversity and determine hard-to-contrast data.
The results of three medical imaging and two natural imaging datasets show that our initial query not only significantly outperforms existing active querying strategies but also surpasses random selection by a large margin. 
This finding is significant because it is the first few choices that define the efficacy and efficiency of the subsequent learning procedure.
We foresee our solution to the cold start problem as a simple, yet strong, baseline to sample the initial query for active learning in image classification.

\noindent{\bf Limitation.} This study provides an empirical benchmark of initial queries in active learning, while more theoretical analyses can be provided. Yehuda~\etal~\cite{yehuda2022active} also found that the choice of active learning strategies depends on the initial query budget. A challenge is to articulate the quantity of determining active learning strategies, which we leave for future work.

\noindent{\bf Potential societal impacts.} Real-world data often exhibit long-tailed distributions, rather than the ideal uniform distributions over each class. We improve active learning by enforcing label diversity and hard-to-contrast data. However, we only extensively test our strategies on academic datasets. In many other real-world domains such as robotics and autonomous driving, the data may impose additional constraints on annotation accessibility or learning dynamics, e.g., being fair or private. We focus on standard accuracy and AUC as our evaluation metrics while ignoring other ethical issues in imbalanced data, especially in underrepresented minority classes.

\section*{Acknowledgements}
\label{sec:acknowledgements}
This work was supported by the Lustgarten Foundation for Pancreatic Cancer Research. The authors want to thank Mingfei Gao for the discussion of initial query quantity and suggestions on the implementation of consistency-based active learning framework.
The authors also want to thank Guy Hacohen, Yuanhan Zhang, Akshay L. Chandra, Jingkang Yang, Hao Cheng, Rongkai Zhang, and Junfei Xiao, for their feedback and constructive suggestions at several stages of the project. Computational resources were provided by Machine Learning and Data Analytics Laboratory, Nanyang Technological University. The authors thank the administrator Sung Kheng Yeo for his technical support.



\bibliographystyle{plain}

{\footnotesize
  \bibliography{refs}

\begin{thebibliography}{10}

\bibitem{Acevedo2020ADO}
Andrea Acevedo, Anna Merino, Santiago Alf{\'e}rez, {\'A}ngel Molina, Laura
  Bold{\'u}, and Jos{\'e} Rodellar.
\newblock A dataset of microscopic peripheral blood cell images for development
  of automatic recognition systems.
\newblock {\em Data in Brief}, 30, 2020.

\bibitem{Agarwal2020ContextualDF}
Sharat Agarwal, Himanshu Arora, Saket Anand, and Chetan Arora.
\newblock Contextual diversity for active learning.
\newblock {\em ArXiv}, abs/2008.05723, 2020.

\bibitem{azizi2021big}
Shekoofeh Azizi, Basil Mustafa, Fiona Ryan, Zachary Beaver, Jan Freyberg,
  Jonathan Deaton, Aaron Loh, Alan Karthikesalingam, Simon Kornblith, Ting
  Chen, et~al.
\newblock Big self-supervised models advance medical image classification.
\newblock {\em arXiv preprint arXiv:2101.05224}, 2021.

\bibitem{balcan2007margin}
Maria-Florina Balcan, Andrei Broder, and Tong Zhang.
\newblock Margin based active learning.
\newblock In {\em International Conference on Computational Learning Theory},
  pages 35--50. Springer, 2007.

\bibitem{bengar2021reducing}
Javad~Zolfaghari Bengar, Joost van~de Weijer, Bartlomiej Twardowski, and Bogdan
  Raducanu.
\newblock Reducing label effort: Self-supervised meets active learning.
\newblock In {\em Proceedings of the IEEE/CVF International Conference on
  Computer Vision}, pages 1631--1639, 2021.

\bibitem{Bengio2009CurriculumL}
Yoshua Bengio, J{\'e}r{\^o}me Louradour, Ronan Collobert, and Jason Weston.
\newblock Curriculum learning.
\newblock In {\em ICML '09}, 2009.

\bibitem{Berman2019MultiGrainAU}
Maxim Berman, Herv{\'e} J{\'e}gou, Andrea Vedaldi, Iasonas Kokkinos, and
  Matthijs Douze.
\newblock Multigrain: a unified image embedding for classes and instances.
\newblock {\em ArXiv}, abs/1902.05509, 2019.

\bibitem{bilic2019liver}
Patrick Bilic, Patrick~Ferdinand Christ, Eugene Vorontsov, Grzegorz Chlebus,
  Hao Chen, Qi~Dou, Chi-Wing Fu, Xiao Han, Pheng-Ann Heng, J{\"u}rgen Hesser,
  et~al.
\newblock The liver tumor segmentation benchmark (lits).
\newblock {\em arXiv preprint arXiv:1901.04056}, 2019.

\bibitem{bobadilla2012collaborative}
Jes{\'u}s Bobadilla, Fernando Ortega, Antonio Hernando, and Jes{\'u}s Bernal.
\newblock A collaborative filtering approach to mitigate the new user cold
  start problem.
\newblock {\em Knowledge-based systems}, 26:225--238, 2012.

\bibitem{borisov2010active}
Alexander Borisov, Eugene Tuv, and George Runger.
\newblock Active batch learning with stochastic query by forest.
\newblock In {\em JMLR: Workshop and Conference Proceedings (2010)}. Citeseer,
  2010.

\bibitem{chandra2021initial}
Akshay~L Chandra, Sai~Vikas Desai, Chaitanya Devaguptapu, and Vineeth~N
  Balasubramanian.
\newblock On initial pools for deep active learning.
\newblock In {\em NeurIPS 2020 Workshop on Pre-registration in Machine
  Learning}, pages 14--32. PMLR, 2021.

\bibitem{chang2017active}
Haw-Shiuan Chang, Erik Learned-Miller, and Andrew McCallum.
\newblock Active bias: Training more accurate neural networks by emphasizing
  high variance samples.
\newblock {\em Advances in Neural Information Processing Systems}, 30, 2017.

\bibitem{chen2020simple}
Ting Chen, Simon Kornblith, Mohammad Norouzi, and Geoffrey Hinton.
\newblock A simple framework for contrastive learning of visual
  representations.
\newblock {\em arXiv preprint arXiv:2002.05709}, 2020.

\bibitem{chen2020improveddemo}
Xinlei Chen, Haoqi Fan, Ross Girshick, and Kaiming He.
\newblock Moco demo: Cifar-10.
\newblock
  \url{https://colab.research.google.com/github/facebookresearch/moco/blob/colab-notebook/colab/moco_cifar10_demo.ipynb}.
\newblock Accessed: 2022-05-26.

\bibitem{chen2020improved}
Xinlei Chen, Haoqi Fan, Ross Girshick, and Kaiming He.
\newblock Improved baselines with momentum contrastive learning.
\newblock {\em arXiv preprint arXiv:2003.04297}, 2020.

\bibitem{farahani2009facility}
Reza~Zanjirani Farahani and Masoud Hekmatfar.
\newblock {\em Facility location: concepts, models, algorithms and case
  studies}.
\newblock Springer Science \& Business Media, 2009.

\bibitem{gal2017deep}
Yarin Gal, Riashat Islam, and Zoubin Ghahramani.
\newblock Deep bayesian active learning with image data.
\newblock In {\em International Conference on Machine Learning}, pages
  1183--1192. PMLR, 2017.

\bibitem{gao2020consistency}
Mingfei Gao, Zizhao Zhang, Guo Yu, Sercan~{\"O} Ar{\i}k, Larry~S Davis, and
  Tomas Pfister.
\newblock Consistency-based semi-supervised active learning: Towards minimizing
  labeling cost.
\newblock In {\em European Conference on Computer Vision}, pages 510--526.
  Springer, 2020.

\bibitem{goyal2019scaling}
Priya Goyal, Dhruv Mahajan, Abhinav Gupta, and Ishan Misra.
\newblock Scaling and benchmarking self-supervised visual representation
  learning.
\newblock In {\em Proceedings of the IEEE International Conference on Computer
  Vision}, pages 6391--6400, 2019.

\bibitem{hacohen2022active}
Guy Hacohen, Avihu Dekel, and Daphna Weinshall.
\newblock Active learning on a budget: Opposite strategies suit high and low
  budgets.
\newblock {\em ArXiv}, abs/2202.02794, 2022.

\bibitem{Hoffer2020AugmentYB}
Elad Hoffer, Tal Ben-Nun, Itay Hubara, Niv Giladi, Torsten Hoefler, and Daniel
  Soudry.
\newblock Augment your batch: Improving generalization through instance
  repetition.
\newblock {\em 2020 IEEE/CVF Conference on Computer Vision and Pattern
  Recognition (CVPR)}, pages 8126--8135, 2020.

\bibitem{holub2008entropy}
Alex Holub, Pietro Perona, and Michael~C Burl.
\newblock Entropy-based active learning for object recognition.
\newblock In {\em 2008 IEEE Computer Society Conference on Computer Vision and
  Pattern Recognition Workshops}, pages 1--8. IEEE, 2008.

\bibitem{houlsby2014cold}
Neil Houlsby, Jos{\'e}~Miguel Hern{\'a}ndez-Lobato, and Zoubin Ghahramani.
\newblock Cold-start active learning with robust ordinal matrix factorization.
\newblock In {\em International conference on machine learning}, pages
  766--774. PMLR, 2014.

\bibitem{huang2021semi}
Siyu Huang, Tianyang Wang, Haoyi Xiong, Jun Huan, and Dejing Dou.
\newblock Semi-supervised active learning with temporal output discrepancy.
\newblock In {\em Proceedings of the IEEE/CVF International Conference on
  Computer Vision}, pages 3447--3456, 2021.

\bibitem{jadon2021covid}
Shruti Jadon.
\newblock Covid-19 detection from scarce chest x-ray image data using few-shot
  deep learning approach.
\newblock In {\em Medical Imaging 2021: Imaging Informatics for Healthcare,
  Research, and Applications}, volume 11601, page 116010X. International
  Society for Optics and Photonics, 2021.

\bibitem{karamcheti2021mind}
Siddharth Karamcheti, Ranjay Krishna, Li~Fei-Fei, and Christopher~D Manning.
\newblock Mind your outliers! investigating the negative impact of outliers on
  active learning for visual question answering.
\newblock {\em arXiv preprint arXiv:2107.02331}, 2021.

\bibitem{Kather2019PredictingSF}
Jakob~Nikolas Kather, Johannes Krisam, Pornpimol Charoentong, Tom Luedde,
  Esther Herpel, Cleo-Aron Weis, Timo Gaiser, Alexander Marx, Nektarios~A.
  Valous, Dyke Ferber, Lina Jansen, Constantino~Carlos Reyes-Aldasoro, Inka
  Z{\"o}rnig, Dirk J{\"a}ger, Hermann Brenner, Jenny Chang-Claude, Michael
  Hoffmeister, and Niels Halama.
\newblock Predicting survival from colorectal cancer histology slides using
  deep learning: A retrospective multicenter study.
\newblock {\em PLoS Medicine}, 16, 2019.

\bibitem{kirsch2019batchbald}
Andreas Kirsch, Joost Van~Amersfoort, and Yarin Gal.
\newblock Batchbald: Efficient and diverse batch acquisition for deep bayesian
  active learning.
\newblock {\em Advances in neural information processing systems}, 32, 2019.

\bibitem{krizhevsky2009learning}
Alex Krizhevsky, Geoffrey Hinton, et~al.
\newblock Learning multiple layers of features from tiny images.
\newblock 2009.

\bibitem{lang2021best}
Adrian Lang, Christoph Mayer, and Radu Timofte.
\newblock Best practices in pool-based active learning for image
  classification.
\newblock 2021.

\bibitem{li2021contrastive}
Yunfan Li, Peng Hu, Zitao Liu, Dezhong Peng, Joey~Tianyi Zhou, and Xi~Peng.
\newblock Contrastive clustering.
\newblock In {\em 2021 AAAI Conference on Artificial Intelligence (AAAI)},
  2021.

\bibitem{litjens2017survey}
Geert Litjens, Thijs Kooi, Babak~Ehteshami Bejnordi, Arnaud Arindra~Adiyoso
  Setio, Francesco Ciompi, Mohsen Ghafoorian, Jeroen~Awm Van Der~Laak, Bram
  Van~Ginneken, and Clara~I S{\'a}nchez.
\newblock A survey on deep learning in medical image analysis.
\newblock {\em Medical image analysis}, 42:60--88, 2017.

\bibitem{margatina2021bayesian}
Katerina Margatina, Loic Barrault, and Nikolaos Aletras.
\newblock Bayesian active learning with pretrained language models.
\newblock {\em arXiv preprint arXiv:2104.08320}, 2021.

\bibitem{mayer2020adversarial}
Christoph Mayer and Radu Timofte.
\newblock Adversarial sampling for active learning.
\newblock In {\em Proceedings of the IEEE/CVF Winter Conference on Applications
  of Computer Vision}, pages 3071--3079, 2020.

\bibitem{mcinnes2018umap}
Leland McInnes, John Healy, and James Melville.
\newblock Umap: Uniform manifold approximation and projection for dimension
  reduction.
\newblock {\em arXiv preprint arXiv:1802.03426}, 2018.

\bibitem{mittal2019parting}
Sudhanshu Mittal, Maxim Tatarchenko, {\"O}zg{\"u}n {\c{C}}i{\c{c}}ek, and
  Thomas Brox.
\newblock Parting with illusions about deep active learning.
\newblock {\em arXiv preprint arXiv:1912.05361}, 2019.

\bibitem{nath2022warm}
Vishwesh Nath, Dong Yang, Holger~R Roth, and Daguang Xu.
\newblock Warm start active learning with proxy labels and selection via
  semi-supervised fine-tuning.
\newblock In {\em International Conference on Medical Image Computing and
  Computer-Assisted Intervention}, pages 297--308. Springer, 2022.

\bibitem{pourahmadi2021simple}
Kossar Pourahmadi, Parsa Nooralinejad, and Hamed Pirsiavash.
\newblock A simple baseline for low-budget active learning.
\newblock {\em arXiv preprint arXiv:2110.12033}, 2021.

\bibitem{qiu2011item}
Tian Qiu, Guang Chen, Zi-Ke Zhang, and Tao Zhou.
\newblock An item-oriented recommendation algorithm on cold-start problem.
\newblock {\em EPL (Europhysics Letters)}, 95(5):58003, 2011.

\bibitem{robinson2020contrastive}
Joshua Robinson, Ching-Yao Chuang, Suvrit Sra, and Stefanie Jegelka.
\newblock Contrastive learning with hard negative samples.
\newblock {\em arXiv preprint arXiv:2010.04592}, 2020.

\bibitem{sener2017active}
Ozan Sener and Silvio Savarese.
\newblock Active learning for convolutional neural networks: A core-set
  approach.
\newblock {\em arXiv preprint arXiv:1708.00489}, 2017.

\bibitem{settles2009active}
Burr Settles.
\newblock Active learning literature survey.
\newblock 2009.

\bibitem{shui2020deep}
Changjian Shui, Fan Zhou, Christian Gagn{\'e}, and Boyu Wang.
\newblock Deep active learning: Unified and principled method for query and
  training.
\newblock In {\em International Conference on Artificial Intelligence and
  Statistics}, pages 1308--1318. PMLR, 2020.

\bibitem{simeoni2021rethinking}
Oriane Sim{\'e}oni, Mateusz Budnik, Yannis Avrithis, and Guillaume Gravier.
\newblock Rethinking deep active learning: Using unlabeled data at model
  training.
\newblock In {\em 2020 25th International Conference on Pattern Recognition
  (ICPR)}, pages 1220--1227. IEEE, 2021.

\bibitem{sinha2019variational}
Samarth Sinha, Sayna Ebrahimi, and Trevor Darrell.
\newblock Variational adversarial active learning.
\newblock In {\em Proceedings of the IEEE/CVF International Conference on
  Computer Vision}, pages 5972--5981, 2019.

\bibitem{sorscher2022beyond}
Ben Sorscher, Robert Geirhos, Shashank Shekhar, Surya Ganguli, and Ari~S
  Morcos.
\newblock Beyond neural scaling laws: beating power law scaling via data
  pruning.
\newblock {\em arXiv preprint arXiv:2206.14486}, 2022.

\bibitem{sourati2019intelligent}
Jamshid Sourati, Ali Gholipour, Jennifer~G Dy, Xavier Tomas-Fernandez, Sila
  Kurugol, and Simon~K Warfield.
\newblock Intelligent labeling based on fisher information for medical image
  segmentation using deep learning.
\newblock {\em IEEE transactions on medical imaging}, 38(11):2642--2653, 2019.

\bibitem{swayamdipta2020dataset}
Swabha Swayamdipta, Roy Schwartz, Nicholas Lourie, Yizhong Wang, Hannaneh
  Hajishirzi, Noah~A Smith, and Yejin Choi.
\newblock Dataset cartography: Mapping and diagnosing datasets with training
  dynamics.
\newblock {\em arXiv preprint arXiv:2009.10795}, 2020.

\bibitem{tong2022videomae}
Zhan Tong, Yibing Song, Jue Wang, and Limin Wang.
\newblock Videomae: Masked autoencoders are data-efficient learners for
  self-supervised video pre-training.
\newblock {\em arXiv preprint arXiv:2203.12602}, 2022.

\bibitem{Touvron2021TrainingDI}
Hugo Touvron, Matthieu Cord, Matthijs Douze, Francisco Massa, Alexandre
  Sablayrolles, and Herv'e J'egou.
\newblock Training data-efficient image transformers \& distillation through
  attention.
\newblock In {\em ICML}, 2021.

\bibitem{van2020scan}
Wouter Van~Gansbeke, Simon Vandenhende, Stamatios Georgoulis, Marc Proesmans,
  and Luc Van~Gool.
\newblock Scan: Learning to classify images without labels.
\newblock In {\em European Conference on Computer Vision}, pages 268--285.
  Springer, 2020.

\bibitem{wang2014new}
Dan Wang and Yi~Shang.
\newblock A new active labeling method for deep learning.
\newblock In {\em 2014 International joint conference on neural networks
  (IJCNN)}, pages 112--119. IEEE, 2014.

\bibitem{medmnistv2}
Jiancheng Yang, Rui Shi, Donglai Wei, Zequan Liu, Lin Zhao, Bilian Ke,
  Hanspeter Pfister, and Bingbing Ni.
\newblock Medmnist v2: A large-scale lightweight benchmark for 2d and 3d
  biomedical image classification.
\newblock {\em arXiv preprint arXiv:2110.14795}, 2021.

\bibitem{yehuda2022active}
Ofer Yehuda, Avihu Dekel, Guy Hacohen, and Daphna Weinshall.
\newblock Active learning through a covering lens.
\newblock {\em arXiv preprint arXiv:2205.11320}, 2022.

\bibitem{yuan2020cold}
Michelle Yuan, Hsuan-Tien Lin, and Jordan Boyd-Graber.
\newblock Cold-start active learning through self-supervised language modeling.
\newblock {\em arXiv preprint arXiv:2010.09535}, 2020.

\bibitem{zhang2010solving}
Zi-Ke Zhang, Chuang Liu, Yi-Cheng Zhang, and Tao Zhou.
\newblock Solving the cold-start problem in recommender systems with social
  tags.
\newblock {\em EPL (Europhysics Letters)}, 92(2):28002, 2010.

\bibitem{zheltonozhskii2020self}
Evgenii Zheltonozhskii, Chaim Baskin, Alex~M Bronstein, and Avi Mendelson.
\newblock Self-supervised learning for large-scale unsupervised image
  clustering.
\newblock {\em arXiv preprint arXiv:2008.10312}, 2020.

\bibitem{zhou2021review}
S~Kevin Zhou, Hayit Greenspan, Christos Davatzikos, James~S Duncan, Bram van
  Ginneken, Anant Madabhushi, Jerry~L Prince, Daniel Rueckert, and Ronald~M
  Summers.
\newblock A review of deep learning in medical imaging: Imaging traits,
  technology trends, case studies with progress highlights, and future
  promises.
\newblock {\em Proceedings of the IEEE}, 2021.

\bibitem{zhou2021towards}
Zongwei Zhou.
\newblock {\em Towards Annotation-Efficient Deep Learning for Computer-Aided
  Diagnosis}.
\newblock PhD thesis, Arizona State University, 2021.

\bibitem{zhou2019integrating}
Zongwei Zhou, Jae Shin, Ruibin Feng, R~Todd Hurst, Christopher~B Kendall, and
  Jianming Liang.
\newblock Integrating active learning and transfer learning for carotid
  intima-media thickness video interpretation.
\newblock {\em Journal of digital imaging}, 32(2):290--299, 2019.

\bibitem{zhou2017fine}
Zongwei Zhou, Jae Shin, Lei Zhang, Suryakanth Gurudu, Michael Gotway, and
  Jianming Liang.
\newblock Fine-tuning convolutional neural networks for biomedical image
  analysis: actively and incrementally.
\newblock In {\em Proceedings of the IEEE Conference on Computer Vision and
  Pattern Recognition}, pages 7340--7349, 2017.

\bibitem{zhou2021active}
Zongwei Zhou, Jae~Y Shin, Suryakanth~R Gurudu, Michael~B Gotway, and Jianming
  Liang.
\newblock Active, continual fine tuning of convolutional neural networks for
  reducing annotation efforts.
\newblock {\em Medical Image Analysis}, page 101997, 2021.

\bibitem{zhu2019addressing}
Yu~Zhu, Jinghao Lin, Shibi He, Beidou Wang, Ziyu Guan, Haifeng Liu, and Deng
  Cai.
\newblock Addressing the item cold-start problem by attribute-driven active
  learning.
\newblock {\em IEEE Transactions on Knowledge and Data Engineering},
  32(4):631--644, 2019.

\end{thebibliography}
}


\clearpage
\appendix

\section{Implementation Configurations}
\label{sec:implementation_appendix}

\subsection{Data Split}
\label{sec:data_split_appendix}

PathMNIST with nine categories has 107,180 colorectal cancer tissue histopathological images extracted from Kather~\etal~\cite{Kather2019PredictingSF}, with 89,996/10,004/7,180 images for training/validation/testing. BloodMNIST contains 17,092 microscopic peripheral blood cell images extracted from Acevedo~\etal~\cite{Acevedo2020ADO} with eight categories, where 11,959/1,712/3,421 images for training/validation/testing. OrganAMNIST consists of the axial view abdominal CT images based on Bilic~\etal~\cite{bilic2019liver}, with 34,581/6,491/17,778 images of 11 categories for training/validation/testing. CIFAR-10-LT ($\rho$=100) consists of a subset of CIFAR-10~\cite{krizhevsky2009learning}, with 12,406/10,000 images for training/testing.

\subsection{Training Recipe for Contrastive Learning}
\label{sec:training_recipe_contrastive_learning_appendix}

\noindent{\bf Pseudocode for Our Proposed Strategy.} The~\algocfautorefname~\ref{alg:initial_selection} provides the pseudocode for our proposed hard-to-contrast initial query strategy, as elaborated in \S\ref{sec:method}.

\begin{algorithm}[!b]
    \caption{\label{alg:initial_selection} Active querying hard-to-contrast data}
    \begin{algorithmic}
        \STATE \textbf{input:}  \\
        \ \ \ $\mathcal{D}=\{\bm x_m\}_{m=1}^M$ \{unlabeled dataset $\mathcal{D}$\ contains $M$ images\}\\
        \ \ \ annotation budget $B$;
        \ the number of clusters $K$;
        \ batch size $N$;
        \ the number of epochs $E$\\
        \ \ \ constant $\tau$; structure of encoder $f$, projection head $g$; augmentation $\mathcal{T}$ \\
        \ \ \ $\theta^{(e)}, e\in [1,E]$ \{model parameters at epoch $e$ during contrastive learning\}
        \STATE \textbf{output:}  \\
        \ \ \ selected query $\mathcal{Q}$ \\

        \STATE $\mathcal{Q} = \varnothing$
        \FOR{epoch $e\in\{1, \ldots, E\}$}
        \FOR{sampled minibatch $\{\bm x_n\}_{n=1}^N$}
        \STATE \textbf{for all} $n\in \{1, \ldots, N\}$ \textbf{do}
        \STATE $~~~~$draw two augmentation functions $t \!\sim\! \mathcal{T}$, $t' \!\sim\! \mathcal{T}$
        \STATE $~~~~$\textcolor{gray}{\# the first augmentation} \STATE $~~~~$$\tilde{\bm x}_{2n-1} = t(\bm x_n)$
                    \STATE $~~~~$$\bm h_{2n-1} = f(\tilde{\bm x}_{2n-1})$  \textcolor{gray}{~~~~~~~~~~~~~~~~~~~~~~~~~~~~~~\# representation}
                \STATE $~~~~$$\bm z_{2n-1} = g({\bm h}_{2n-1})$  \textcolor{gray}{~~~~~~~~~~~~~~~~~~~~~~~~~~~~~~~\# projection}
                    \STATE $~~~~$\textcolor{gray}{\# the second augmentation} \STATE $~~~~$$\tilde{\bm x}_{2n} = t'(\bm x_n)$
                \STATE $~~~~$$\bm h_{2n} = f(\tilde{\bm x}_{2n})$      \textcolor{gray}{~~~~~~~~~~~~~~~~~~~~~~~~~~~~~~~~~~~~~~\# representation}
                    \STATE $~~~~$$\bm z_{2n} = g({\bm h}_{2n})$      \textcolor{gray}{~~~~~~~~~~~~~~~~~~~~~~~~~~~~~~~~~~~~~~~\# projection}
                \STATE \textbf{end for}
                \STATE \textbf{for all} $i\in\{1, \ldots, 2N\}$ and $j\in\{1, \dots, 2N\}$ \textbf{do}
                \STATE $~~~~$ $s_{i,j} = \bm z_i^\top \bm z_j / (\lVert\bm z_i\rVert \lVert\bm z_j\rVert)$ \textcolor{gray}{~~~~~~~~~~~\# pairwise similarity}\\
                \STATE $~~~~$ $p_{i,j} = \frac{\exp(s_{i,j})/\tau}{\sum_{n=1}^{2N}\one{n \neq i}\exp(s_{i,n})/\tau}$ \textcolor{gray}{~~~~~\# predicted probability of contrastive pre-text task}\\
                \STATE \textbf{end for}
                \STATE  $p_{\theta^{(e)}}(y_n^*|x_n)=\frac{1}{2}[p_{2n-1, 2n}+p_{2n, 2n-1}]$ \\
                \ENDFOR
                \ENDFOR
                \FOR{unlabeled images $\{\bm x_m\}_{m=1}^M$}
                \STATE $\hat{\mu}_m = \frac{1}{E}\sum_{e=1}^{E}p_{\theta^{(e)}}(y_m^*|x_m)$ \\
                \STATE Assign $\bm x_m$ to one of the clusters computed by $K$-mean($\bm h$, $K$)\\
                \ENDFOR
                \STATE \textbf{for all} $k\in \{1, \ldots, K\}$ \textbf{do}
                \STATE $~~~~$ sort images in the cluster $K$ based on $\hat{\mu}$ in an ascending order
                \STATE $~~~~$ query labels for top $B/K$ samples, yielding ${Q}_k$
                \STATE $~~~~$ $\mathcal{Q} = \mathcal{Q} \cup \mathcal{Q}_k$
                \STATE \textbf{end for}
                \STATE \textbf{return} $\mathcal{Q}$
    \end{algorithmic}
\end{algorithm}
\noindent{\bf Pre-training Settings.} Our settings mostly follow~\cite{chen2020improved, chen2020improveddemo}.~\tableautorefname~\ref{tab:contrastive_detail_medmnist} summarizes our contrastive pre-training settings on MedMNIST, following~\cite{chen2020improved}.~\tableautorefname~\ref{tab:contrastive_detail_medmnist} shows the corresponding pre-training settings on CIFAR-10-LT, following the official MoCo demo on CIFAR-10~\cite{chen2020improveddemo}. The contrastive learning model is pre-trained on 2 NVIDIA RTX3090 GPUs with 24GB memory each. The total number of model parameters is 55.93 million, among which 27.97 million requires gradient backpropagation.

\begin{table}[t]
\centering
    \caption{Contrastive learning settings on MedMNIST and CIFAR-10-LT.}
    \subfloat[{MedMNIST pre-training}\label{tab:contrastive_detail_medmnist}]{
        \tablestyle{3pt}{1.00}
        \begin{tabular}{y{85}|x{75}}
            config                                       & value                                               \\
            \shline
            backbone                                     & {ResNet-50}                                         \\
            optimizer                                    & {SGD}                                               \\
            optimizer momentum                           & {0.9}                                               \\
            weight decay                                 & {1e-4}                                              \\
            base learning rate$^\dagger$                 & 0.03                                                \\
            learning rate schedule                       & {cosine decay}                                      \\
            warmup epochs                                & {5}                                                 \\
            epochs                                       & {200}                                               \\
            repeated sampling~\cite{Hoffer2020AugmentYB} & see~\tableautorefname~\ref{tab:repeat_augmentation} \\
            \multirow{1}{*}{augmentation}                & see~\tableautorefname~\ref{tab:image_augmentation}  \\
            batch size                                   & 4096                                                \\
            queue length~\cite{chen2020improved}         & 65536                                               \\
            $\tau$~(\eqref{eq:p_ij})                & 0.05                                                \\
        \end{tabular}
    }
    \hspace{30pt}
    \subfloat[{CIFAR-10-LT pre-training}\label{tab:contrastive_detail_cifar}]{
        \tablestyle{3pt}{1.00}
        \begin{tabular}{y{85}|x{75}}
            config                                       & value                                              \\
            \shline
            backbone                                     & {ResNet-50}                                        \\
            optimizer                                    & {SGD}                                              \\
            optimizer momentum                           & {0.9}                                              \\
            weight decay                                 & {1e-4}                                             \\
            base learning rate$^\dagger$                 & 0.03                                               \\
            learning rate schedule                       & {cosine decay}                                     \\
            warmup epochs                                & {5}                                                \\
            epochs                                       & {800}                                              \\
            repeated sampling~\cite{Hoffer2020AugmentYB} & none                                               \\
            \multirow{1}{*}{augmentation}                & see~\tableautorefname~\ref{tab:image_augmentation} \\
            batch size                                   & 512                                                \\
            queue length~\cite{chen2020improved}         & 4096                                               \\
            $\tau$~(\eqref{eq:p_ij})                & 0.05                                               \\
        \end{tabular}
    }
    \begin{tablenotes}
        \scriptsize
        \item $^\dagger$\textit{lr} = \textit{base\_lr}$\times$batchsize / 256 per the linear \textit{lr} scaling rule~\cite{goyal2019scaling}.
    \end{tablenotes}
\end{table}

\noindent{\bf Dataset Augmentation.} We apply the same augmentation as in MoCo v2~\cite{chen2020improved} on all the images of RGB modalities to reproduce the optimal augmentation pipeline proposed by the authors, including PathMNIST, BloodMNIST, CIFAR-10-LT. Because OrganAMNIST is a grey scale CT image dataset, we apply the augmentation in~\cite{azizi2021big} designed for radiological images, replacing random gray scale and Gaussian blur with random rotation.~\tableautorefname~\ref{tab:image_augmentation} shows the details of data augmentation.


\begin{table}[t]\centering
    \caption{\textbf{Data augmentations.}}
    \subfloat[{Augmentations for RGB images}\label{tab:augmentation_color}]{
        \tablestyle{3pt}{1.00}
        \begin{tabular}{y{55}|x{105}}
            augmentation  & value                                           \\
            \shline
            hflip         &                                                 \\
            crop          & {[0.08, 1]}                                     \\
            color jitter  & {[0.4, 0.4, 0.4, 0.1], p=0.8}                   \\
            gray scale    &                                                 \\
            Gaussian blur & {$\sigma_{min}$=0.1, $\sigma_{max}$=2.0, p=0.5} \\
        \end{tabular}
    }
    \hspace{15pt}
    \subfloat[{Augmentations for OrganAMNIST}\label{tab:augmentation_bw}]{
        \tablestyle{3pt}{1.00}
        \begin{tabular}{y{55}|x{105}}
            augmentation & value                         \\
            \shline
            hflip        &                               \\
            crop         & {[0.08, 1]}                   \\
            color jitter & {[0.4, 0.4, 0.4, 0.1], p=0.8} \\
            rotation     & {degrees=45}                  \\
            \multicolumn{2}{c}{~}                        \\
        \end{tabular}
    }
    \label{tab:image_augmentation}
\end{table}

\noindent{\bf Repeated Augmentation.} Our MoCo v2 pre-training is so fast in computation that data loading becomes a new bottleneck that dominates running time in our setup. We perform repeated augmentation on MedMNIST datasets at the level of dataset, also to enlarge augmentation space and improve generalization.~\cite{Hoffer2020AugmentYB} proposed repeated augmentation in a growing batch mode to improve generalization and convergence speed by reducing variances. This approach provokes a challenge in computing resources. Recent works \cite{Hoffer2020AugmentYB, Touvron2021TrainingDI, Berman2019MultiGrainAU} proved that fixed batch mode also boosts generalization and optimization by increasing mutiplicity of augmentations as well as parameter updates and decreasing the number of unique samples per batch, which holds the batch size fixed. Because the original contrastive learning works~\cite{chen2020simple,chen2020improved} were implemented on ImageNet dataset, we attempt to simulate the quantity of ImageNet per epoch to achieve optimal performances. The details are shown in~\tableautorefname~\ref{tab:repeat_augmentation}.

We only applied repeated augmentation on MedMNIST, but not CIFAR-10-LT. This is because we follow all the settings of the official CIFAR-10 demo~\cite{chen2020improveddemo} in which repeated augmentation is not employed.

\begin{table*}[t]
\footnotesize
    \centering
    \caption{\textbf{Repeated augmentation.} For a faster model convergence, we apply repeated augmentation~\cite{Hoffer2020AugmentYB,tong2022videomae,Touvron2021TrainingDI} on MedMNIST by reproducing the large batch size and iteration numbers.
    }
    \label{tab:repeat_augmentation}
    \begin{tabular}{p{0.23\linewidth}P{0.2\linewidth}P{0.15\linewidth}P{0.3\linewidth}}
                                & \# training & repeated times & \# samples per epoch \\
        \shline
        ImageNet                & 1,281,167   & 1              & 1,281,167            \\
        \hline
        PathMNIST               & 89,996      & 14             & 1,259,944            \\
        OrganAMNIST             & 34,581      & 37             & 1,279,497            \\
        BloodMNIST              & 11,959      & 105            & 1,255,695            \\
        \hline
        CIFAR-10-LT($\rho$=100) & 12,406      & 1              & 12,406               \\
        \hline
    \end{tabular}
\end{table*}

\subsection{Training Recipe for MedMNIST and CIFAR-10}
\label{sec:training_recipe_medmnist_appendix}

\noindent{\bf Benchmark Settings.} We evaluate the initial queries by the performance of model trained on the selected initial query, and present the results in~\tableautorefname~\ref{tab:label_diversity},~\ref{tab:label_diversity_cifar10lt} and~\figureautorefname~\ref{fig:histogram}. The benchmark experiments are performed on NVIDIA RTX 1080 GPUs, with the following settings in~\tableautorefname~\ref{tab:benchmark_settings}.

\noindent{\bf Cold Start Settings for Existing Active Querying Criteria.} To compare the cold start performance of active querying criteria with random selection (~\figureautorefname~\ref{fig:teaser_cold_start}), we trained a model with the test set and applied existing active querying criteria.

\begin{table}[!h]\centering
    \tablestyle{3pt}{1.00}
    \caption{\textbf{Benchmark settings.} We apply the same settings for training MedMNIST, CIFAR-10, and CIFAR-10-LT.}
    \begin{tabular}{y{85}|x{250}}
        config                        & value                                                       \\
        \shline
        backbone                      & {Inception-ResNet-v2}                                       \\
        optimizer                     & {SGD}                                                       \\
        learning rate                 & {0.1}                                                       \\
        learning rate schedule        & reduce learning rate on plateau, factor=0.5, patience=8     \\
        early stopping patience       & {50}                                                        \\
        max epochs                    & {10000}                                                     \\
        \multirow{4}{*}{augmentation} & flip, p=0.5                                                 \\
                                      & rotation, p=0.5, in 90, 180, or 270 degrees                 \\
                                      & reverse color, p=0.1                                        \\
                                      & fade color, p=0.1, 80\% random noises + 20\% original image \\
        batch size                    & {128}                                                       \\
    \end{tabular}
    \label{tab:benchmark_settings}
\end{table}

\clearpage
\section{Additional Results on MedMNIST}

\subsection{Label Diversity is a Significant Add-on to Most Querying Strategies}

\begin{figure}[!h]
    \includegraphics[width=\linewidth]{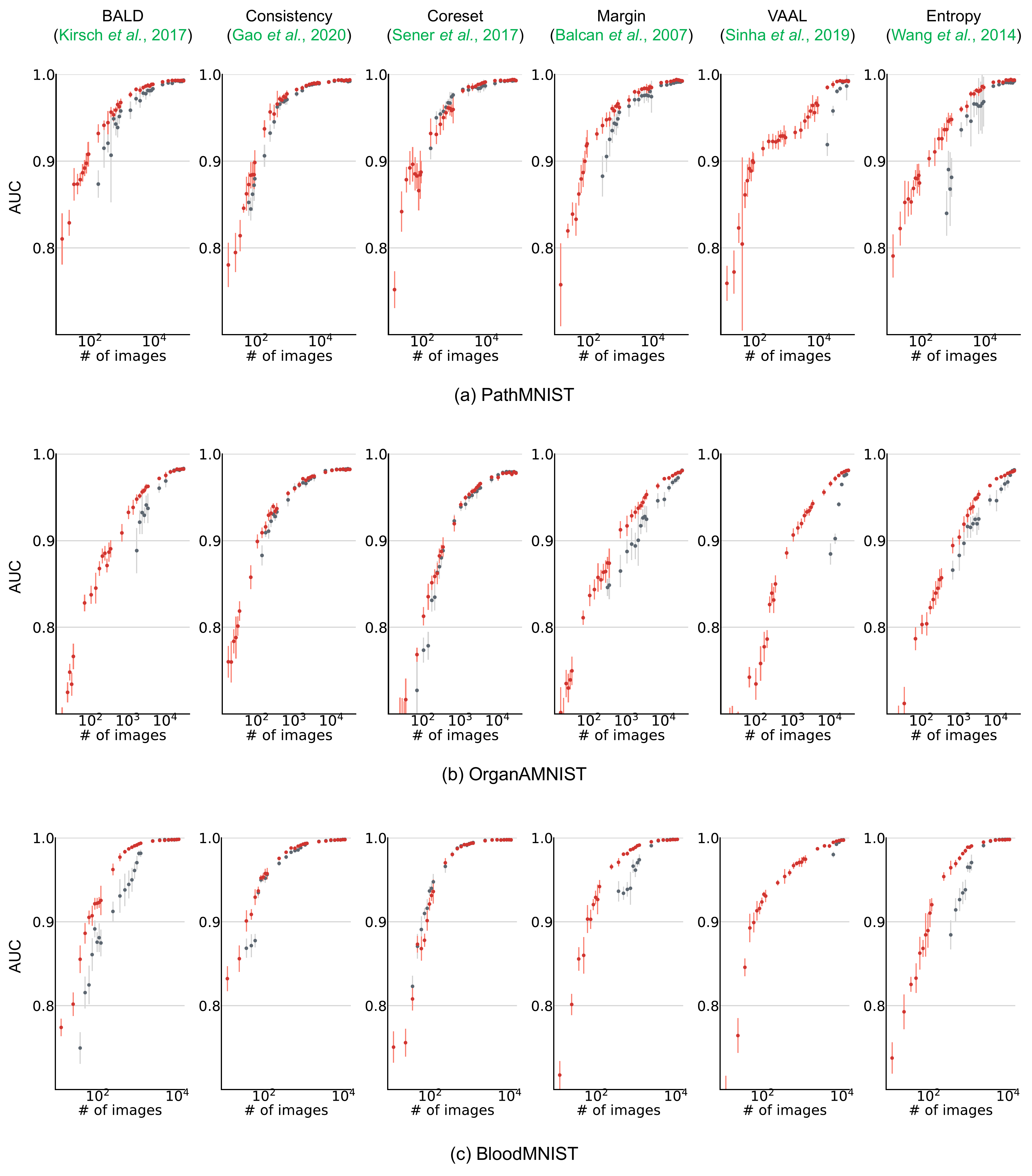}
    \caption{
        [Extended from \tableautorefname~\ref{tab:label_diversity}] \textbf{Label diversity yields more performant and robust active querying strategies.} The experiments are conducted on three datasets in MedMNIST. The red and gray dots denote AUC scores of different active querying strategies with and without label diversity, respectively.
        Most existing active querying strategies became more performant and robust in the presence of label diversity, \eg BALD, Margin, VAAL, and Uncertainty in particular.
        Some gray dots are not plotted in the low budget regime because there are classes absent in the queries due to the selection bias.
    }
    \label{fig:label_diversity_appendix}
\end{figure}

As we present in~\tableautorefname~\ref{tab:label_diversity}, label diversity is an important underlying criterion in designing active querying criteria. We plot the full results on all three MedMNIST datasets in~\figureautorefname~\ref{fig:label_diversity_appendix}. Most existing active querying strategies became more performant and robust in the presence of label diversity.

\subsection{Contrastive Features Enable Label Diversity to Mitigate Bias}

Our proposed active querying strategy is capable of covering the majority of classes in most low budget scenarios by integrating K-means clustering and contrastive features, including the tail classes (\eg femur-left, basophil). Compared to the existing active querying criteria, we achieve the best class coverage of selected query among at all budgets presented in~\tableautorefname~\ref{tab:label_coverage}.

\begin{figure}[!h]
    \includegraphics[width=\linewidth]{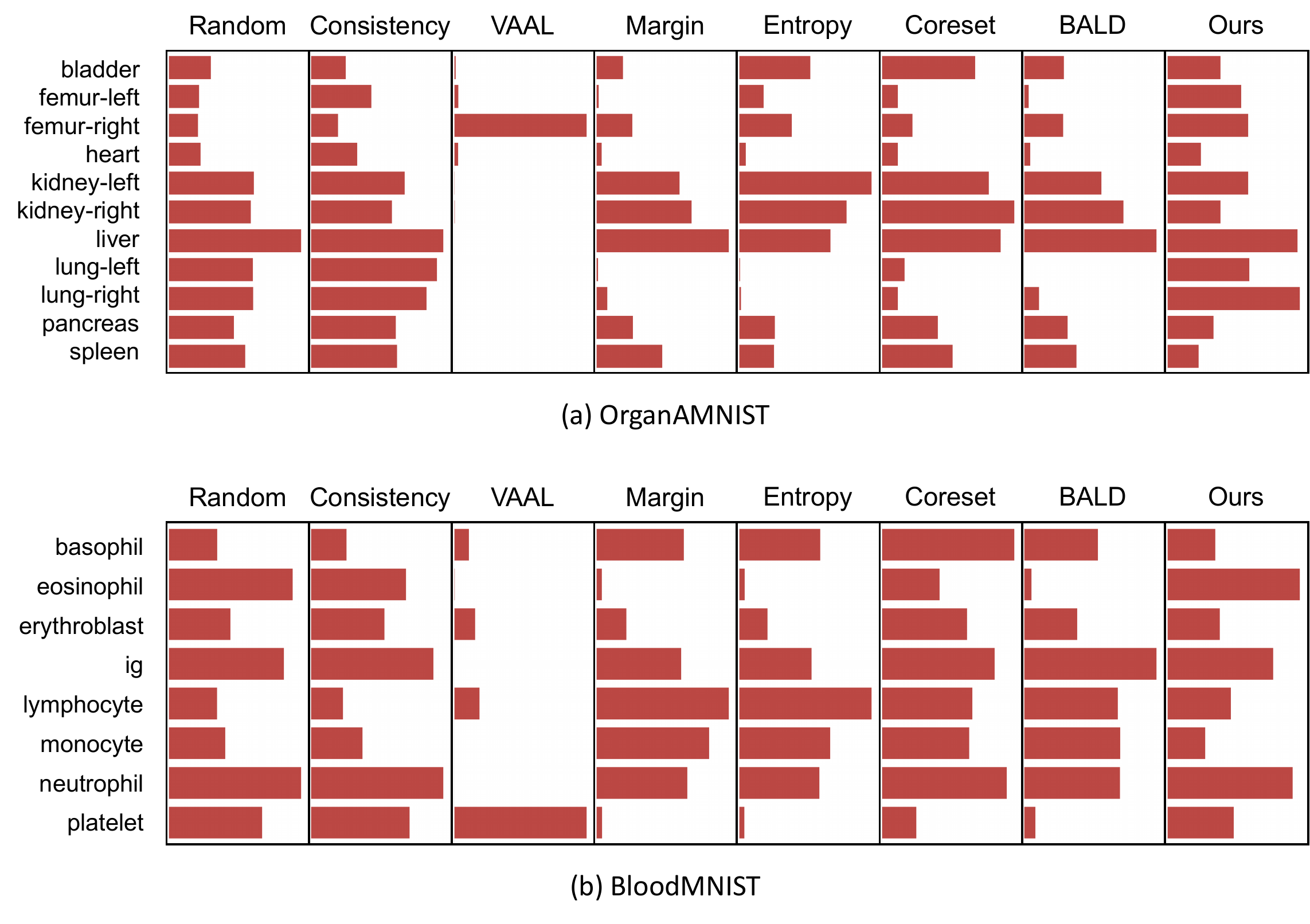}
    \caption{
        [Continued from \figureautorefname~\ref{fig:selected_label_histogram_pathmnist}] \textbf{Our querying strategy yields better label diversity.}
        Random on the leftmost denotes the class distribution of randomly queried samples, which can also reflect the approximate class distribution of the entire dataset.
        As seen, even with a relatively larger initial query budget (691 images, 2\% of OrganAMNIST, and 2,391 images, 20\% of BloodMNIST), most active querying strategies are biased towards certain classes.
        For example in OrganAMNIST, VAAL prefers selecting data in the femur-right and platelet class, but largely ignores data in the lung, liver and monocyte classes.
        On the contrary, our querying strategy not only selects more data from minority classes (e.g., femur-left and basophil) while retaining the class distribution of major classes.
    }
    \label{fig:selected_label_histogram_appendix}
\end{figure}

\clearpage

\begin{figure}[!t]
    \includegraphics[width=\linewidth]{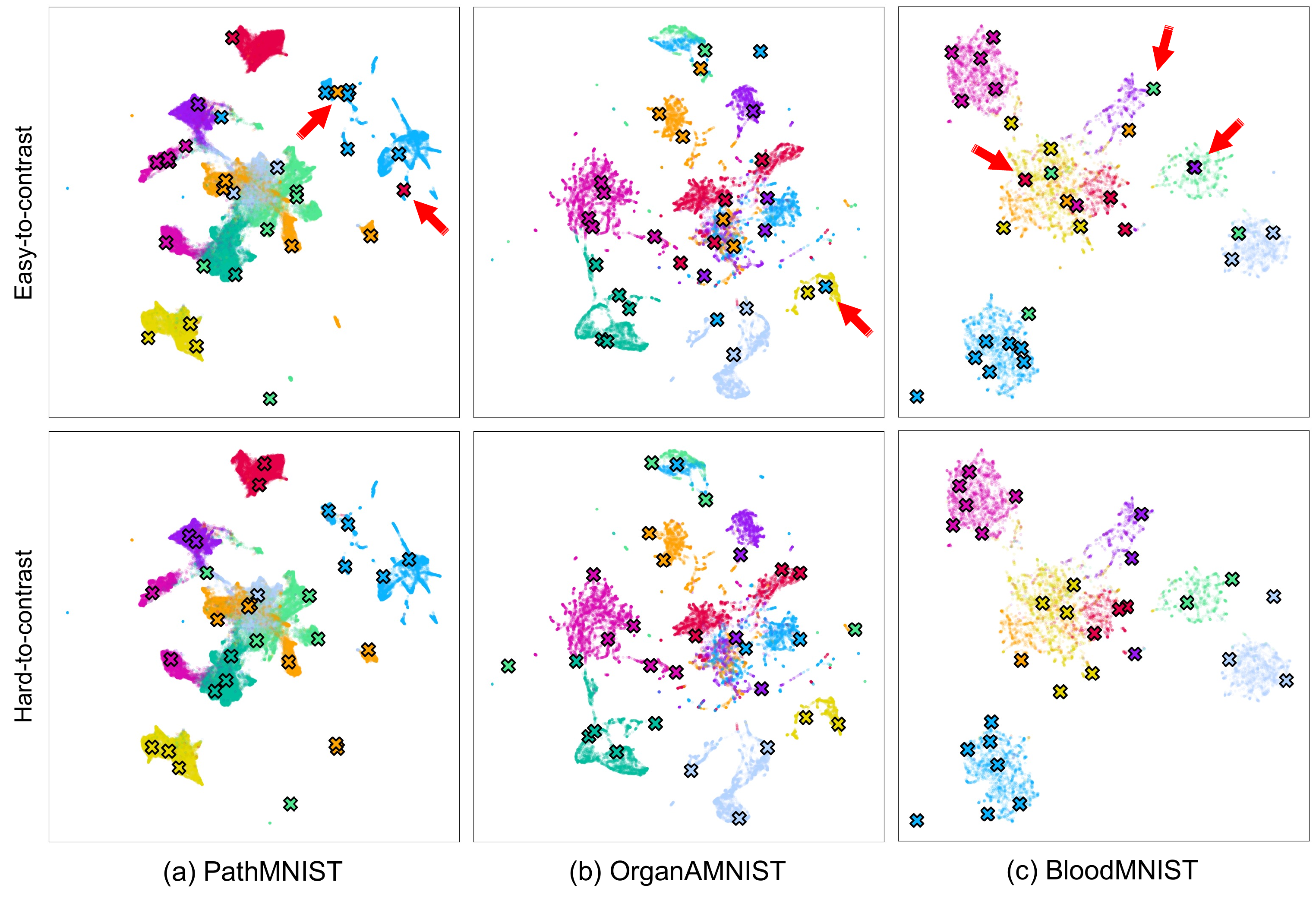}
    \caption{
        \textbf{Visualization of $K$-means clustering and our active selection.}
        UMAP~\cite{mcinnes2018umap} is used to visualize the feature clustering.
        Colors indicate the ground truth.
        Contrastive features clustered by the $K$-means algorithm present a fairly clear separation in the 2D space, which helps enforce the label diversity without the need of ground truth.
        The crosses denote the selected easy- (top) and hard-to-contrast (bottom) data.
        Overall, hard-to-contrast data have a greater spread within each cluster than easy-to-contrast ones.
        In addition, we find that easy-to-contrast tends to select outlier classes that do not belong to the majority class in a cluster (see red arrows). This behavior will invalidate the purpose of clustering and inevitably jeopardize the label diversity.
    }
    \label{fig:umap}
\end{figure}

\noindent\textbf{Selected Query Visualization.}
To ease the analysis, we project the image features (extracted by a trained MoCo v2 encoder) onto a 2D space by UMAP~\cite{mcinnes2018umap}.
The assigned pseudo labels have large overlap with ground truths, suggesting that the features from MoCo v2 are quite discriminative for each class.
Overall, \figureautorefname~\ref{fig:umap} shows that hard-to-contrast queries have a greater spread within each cluster than easy-to-contrast ones.
Both strategies can cover 100\% classes.
Nevertheless, we notice that easy-to-contrast selects \textit{local outliers} in clusters: samples that do not belong to the majority class in a cluster. Such behavior will invalidate the purpose of clustering, which is to query uniformly by separating classes. Additionally, it possibly exposes the risk of introducing out-of-distribution data to the query, which undermines active learning~\cite{karamcheti2021mind}.

\clearpage
\section{Experiments on CIFAR-10 and CIFAR-10-LT}
\label{sec:cifar10lt_appendix}

\subsection{Label Diversity is a Significant Add-on to Most Querying Strategies}

As illustrated in~\tableautorefname~\ref{tab:label_diversity_cifar10lt} and~\figureautorefname~\ref{fig:label_diversity_cifar_appendix}, label diversity is an important underlying criterion in designing active querying criteria on CIFAR-10-LT, an extremely imbalanced dataset. We compare the results of CIFAR-10-LT with MedMNIST datasets~\figureautorefname~\ref{fig:label_diversity_appendix}. CIFAR-10-LT is more imbalanced than MedMNIST, and the performance gain and robustness improvement of label diversity CIFAR-10-LT is significantly larger than MedMNIST. Most of the active querying strategies fail to query all the classes even at relatively larger initial query budgets.

\begin{table*}[h]
    \footnotesize
    \centering
    \caption{
        \textbf{Diversity is a significant add-on to most querying strategies.}
        AUC scores of different querying strategies are compared on CIFAR-10 and CIFAR-10-LT.
        In the low budget regime (\eg 10\% and 20\% of the entire dataset), active querying strategies benefit from enforcing the label diversity of the selected data.
        The cells are highlighted in blue when adding diversity performs no worse than the original querying strategies.
        Some results are missing (marked as ``-'') because the querying strategy fails to sample at least one data point for each class.
        Results of more sampling ratios are presented in Appendix~\figureautorefname~\ref{fig:label_diversity_cifar_appendix}.
    }
    \label{tab:label_diversity_cifar10lt}
    \begin{tabular}{p{0.1\linewidth}P{0.06\linewidth}|P{0.1\linewidth}P{0.1\linewidth}P{0.1\linewidth}P{0.1\linewidth}P{0.1\linewidth}P{0.1\linewidth}@{}}
        \toprule
                                     &         & \multicolumn{6}{c}{CIFAR-10-LT}                                                                                                                                                                                 \\
                                     &         & 1\%                              & 5\%                              & 10\%                             & 20\%                             & 30\%                             & 40\%                             \\
                                     & Unif.   & (142)                            & (710)                            & (1420)                           & (2841)                           & (4261)                           & (5682)                           \\
        \shline
        \multirow{2}{*}{Consistency} & \cmark  & \cellcolor{iblue!30}78.0$\pm$1.2 & \cellcolor{iblue!30}90.0$\pm$0.1 & \cellcolor{iblue!30}91.4$\pm$1.1 & \cellcolor{iblue!30}93.4$\pm$0.2 & \cellcolor{iblue!30}93.2$\pm$0.2 & \cellcolor{iblue!30}94.6$\pm$0.2 \\
                                     & \xmarkg & -                                & -                                & 67.1$\pm$17.1                    & 88.6$\pm$0.3                     & 90.4$\pm$0.6                     & 90.7$\pm$0.2                     \\
        \hline
        \multirow{2}{*}{VAAL}        & \cmark  & \cellcolor{iblue!30}80.9$\pm$1.0 & \cellcolor{iblue!30}90.3$\pm$0.5 & \cellcolor{iblue!30}92.6$\pm$0.2 & \cellcolor{iblue!30}93.7$\pm$0.4 & \cellcolor{iblue!30}93.9$\pm$0.8 & \cellcolor{iblue!30}94.5$\pm$0.2 \\
                                     & \xmarkg & -                                & -                                & -                                & -                                & -                                & 77.3$\pm$1.6                     \\
        \hline
        \multirow{2}{*}{Margin}      & \cmark  & \cellcolor{iblue!30}81.2$\pm$1.8 & \cellcolor{iblue!30}88.7$\pm$0.7 & \cellcolor{iblue!30}91.7$\pm$0.9 & \cellcolor{iblue!30}93.2$\pm$0.2 & \cellcolor{iblue!30}94.5$\pm$0.1 & \cellcolor{iblue!30}94.7$\pm$0.4 \\
                                     & \xmarkg & -                                & -                                & 81.9$\pm$0.8                     & 86.3$\pm$0.3                     & 87.4$\pm$0.2                     & 88.1$\pm$0.1                     \\
        \hline
        \multirow{2}{*}{Entropy}     & \cmark  & \cellcolor{iblue!30}78.1$\pm$1.4 & \cellcolor{iblue!30}89.6$\pm$0.5 & \cellcolor{iblue!30}92.0$\pm$1.2 & \cellcolor{iblue!30}91.9$\pm$1.3 & \cellcolor{iblue!30}94.0$\pm$0.6 & \cellcolor{iblue!30}94.0$\pm$0.7 \\
                                     & \xmarkg & -                                & 79.0$\pm$1.2                     & 65.6$\pm$15.6                    & 86.4$\pm$0.2                     & 88.5$\pm$0.2                     & 89.5$\pm$0.7                     \\
        \hline
        \multirow{2}{*}{Coreset}     & \cmark  & \cellcolor{iblue!30}80.8$\pm$1.0 & \cellcolor{iblue!30}89.7$\pm$1.3 & \cellcolor{iblue!30}91.5$\pm$0.4 & \cellcolor{iblue!30}93.6$\pm$0.2 & \cellcolor{iblue!30}93.4$\pm$0.7 & \cellcolor{iblue!30}94.8$\pm$0.1 \\
                                     & \xmarkg & -                                & -                                & 65.9$\pm$15.9                    & 86.9$\pm$0.1                     & 88.2$\pm$0.1                     & 90.3$\pm$0.2                     \\
        \hline
        \multirow{2}{*}{BALD}        & \cmark  & \cellcolor{iblue!30}83.3$\pm$0.6 & \cellcolor{iblue!30}90.8$\pm$0.3 & \cellcolor{iblue!30}92.8$\pm$0.1 & \cellcolor{iblue!30}90.8$\pm$2.4 & \cellcolor{iblue!30}94.0$\pm$0.8 & \cellcolor{iblue!30}94.7$\pm$0.4 \\
                                     & \xmarkg & -                                & 76.8$\pm$2.3                     & 64.9$\pm$14.9                    & 84.7$\pm$0.6                     & 88.0$\pm$0.5                     & 88.9$\pm$0.1                     \\
        \bottomrule
    \end{tabular}
\end{table*}

\subsection{Contrastive Features Enable Label Diversity to Mitigate Bias}

\begin{figure}[!h]
    \includegraphics[width=\linewidth]{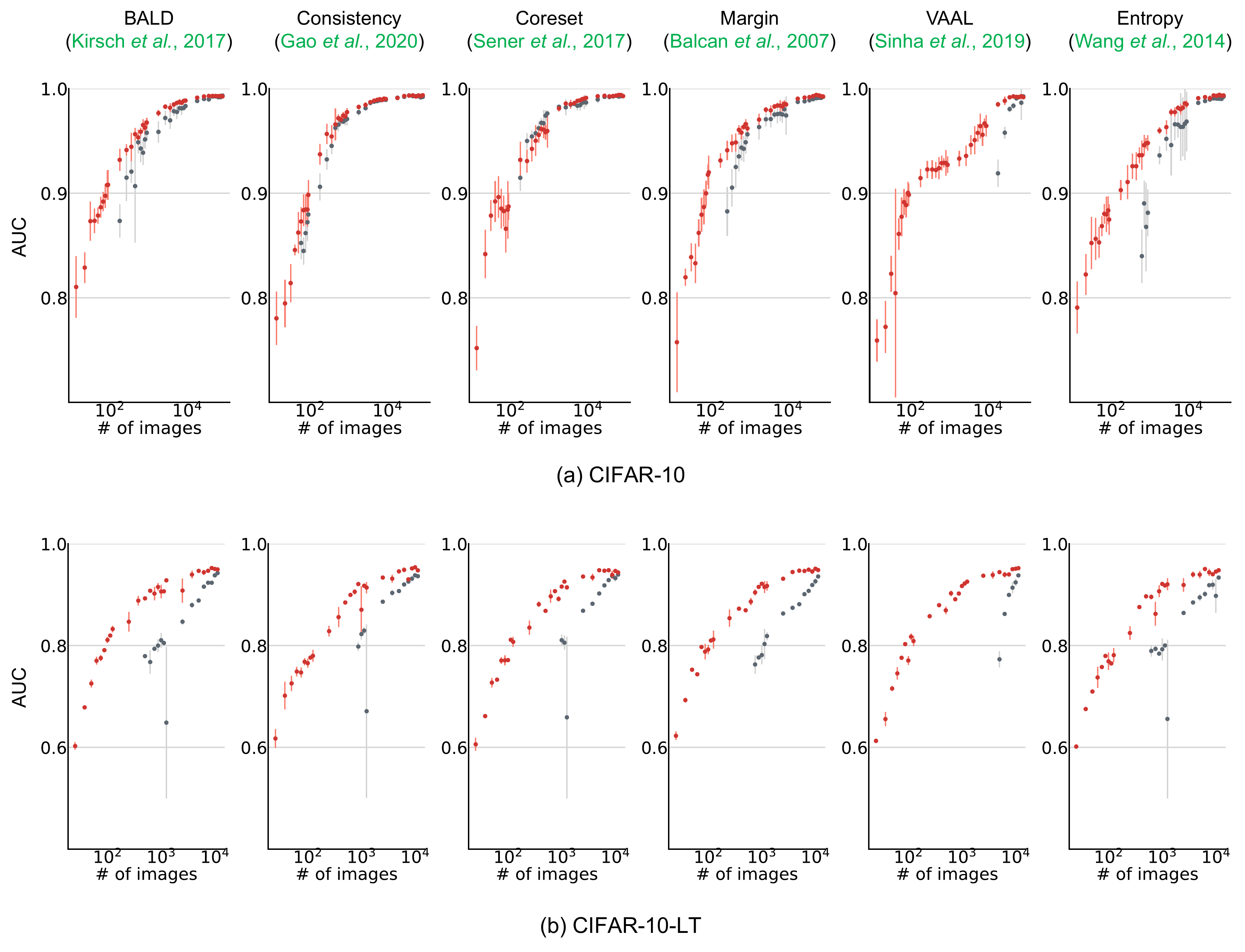}
    \caption{\textbf{Diversity yields more performant and robust active querying strategies.} The experiments are conducted on CIFAR-10-LT. The red and gray dots denote AUC scores of different active querying strategies with and without label diversity, respectively.
        Observations are consistent with those in medical applications (see \figureautorefname~\ref{fig:label_diversity_appendix}):
        Most existing active querying strategies became more performant and robust in the presence of label diversity.
    }
    \label{fig:label_diversity_cifar_appendix}
\end{figure}

Our proposed active querying strategy is capable of covering the majority of classes in most low budget scenarios by integrating K-means clustering and contrastive features, including the tail classes (horse, ship, and truck). Compared to the existing active querying criteria, we achieve the best class coverage of selected query among at all budgets presented in~\tableautorefname~\ref{tab:label_coverage}. As depicted in~\figureautorefname~\ref{fig:label_diversity_cifar_appendix}, our querying strategy has a more similar distribution to the overall distribution of dataset and successfully covers all the classes, with the highest proportion of minor classes (ship and truch) among random selection and all active querying methods.

\begin{figure}[t]
    \includegraphics[width=\linewidth]{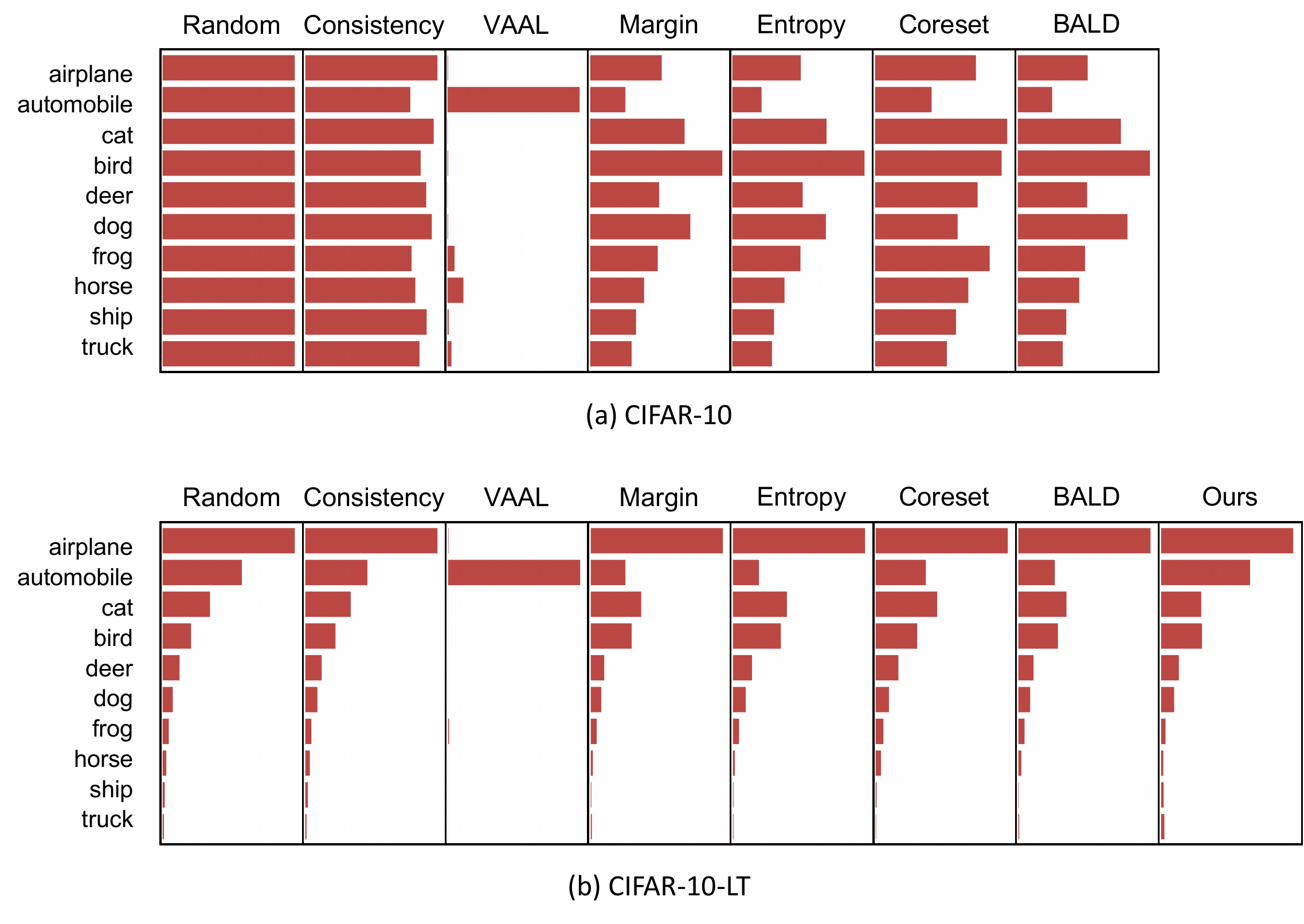}
    \caption{
        \textbf{Our querying strategy yields better label diversity.}
        Random on the leftmost denotes the class distribution of randomly queried samples, which can also reflect the approximate class distribution of the entire dataset.
        As seen, even with a relatively larger initial query budget (5,000 images, 10\% of CIFAR-10, and 1420 images, 10\% of CIFAR-10-LT), most active querying strategies are biased towards certain classes.
        Our querying strategy, on the contrary, is capable of selecting more data from the minority classes such as horse, ship, and truck.
    }
    \label{fig:selected_label_histogram_cifar_appendix}
\end{figure}


\begin{figure}[!t]
    \includegraphics[width=1.0\linewidth]{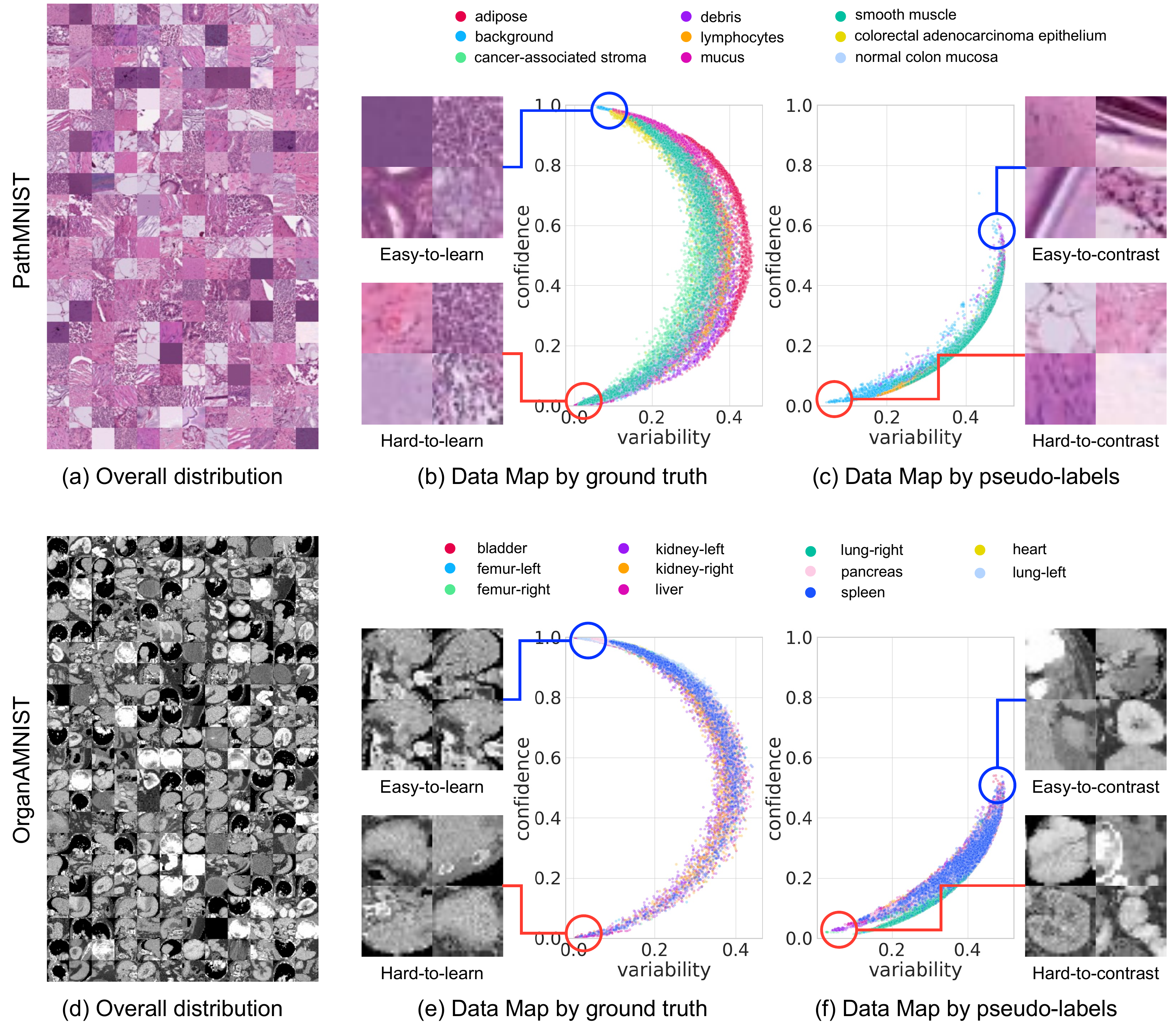}
    \caption{\textbf{Active querying based on Dataset Maps.}
        (a,d) PathMNIST and OrganAMNIST dataset overview.
        (b,e) Easy- and hard-to-learn data can be selected from the maps based on ground truths~\cite{karamcheti2021mind}. This querying strategy has two limitations: (1) requiring manual annotations and (2) data are stratified by classes in the 2D space, leading to a poor label diversity in the selected queries.
        (c,f) Easy- and hard-to-contrast data can be selected from the maps based on pseudo labels.
        This querying strategy is label-free and the selected ``hard-to-contrast'' data represent the most common patterns in the entire dataset. These data are more suitable for training and thus alleviate the cold start problem.
    }
    \label{fig:cartography_path_organa}
\end{figure}

\begin{figure}[!t]
    \includegraphics[width=1.0\linewidth]{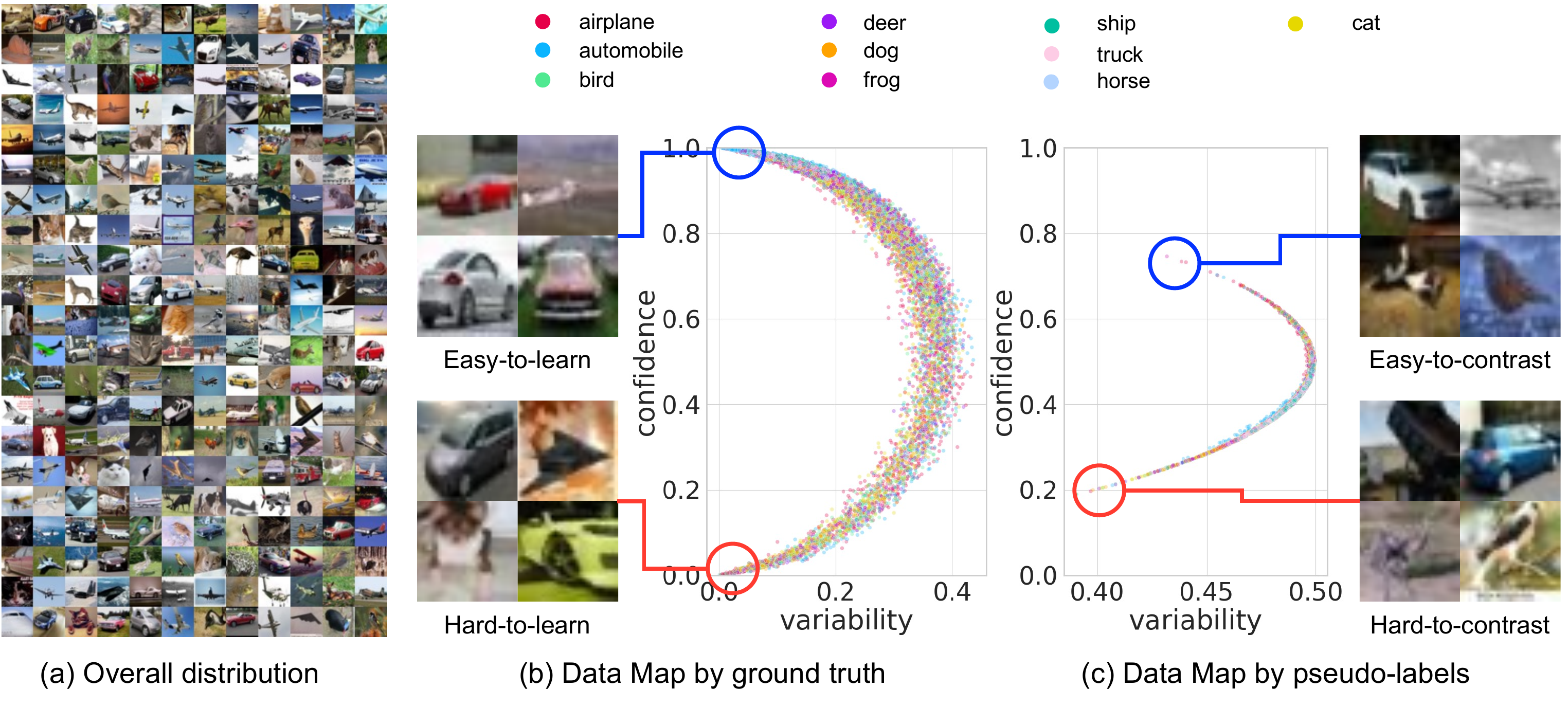}
    \caption{\textbf{Active querying based on Dataset Maps.}
        (a) CIFAR-10-LT dataset overview.
        (b) Easy- and hard-to-learn data can be selected from the maps based on ground truths~\cite{karamcheti2021mind}. This querying strategy has two limitations: (1) requiring manual annotations and (2) data are stratified by classes in the 2D space, leading to a poor label diversity in the selected queries.
        (c) Easy- and hard-to-contrast data can be selected from the maps based on pseudo labels.
        This querying strategy is label-free and the selected ``hard-to-contrast'' data represent the most common patterns in the entire dataset. These data are more suitable for training and thus alleviate the cold start problem.
    }
    \label{fig:cartography_cifar10lt_appendix}
\end{figure}

\clearpage

\begin{figure}[!t]
    \includegraphics[width=\linewidth]{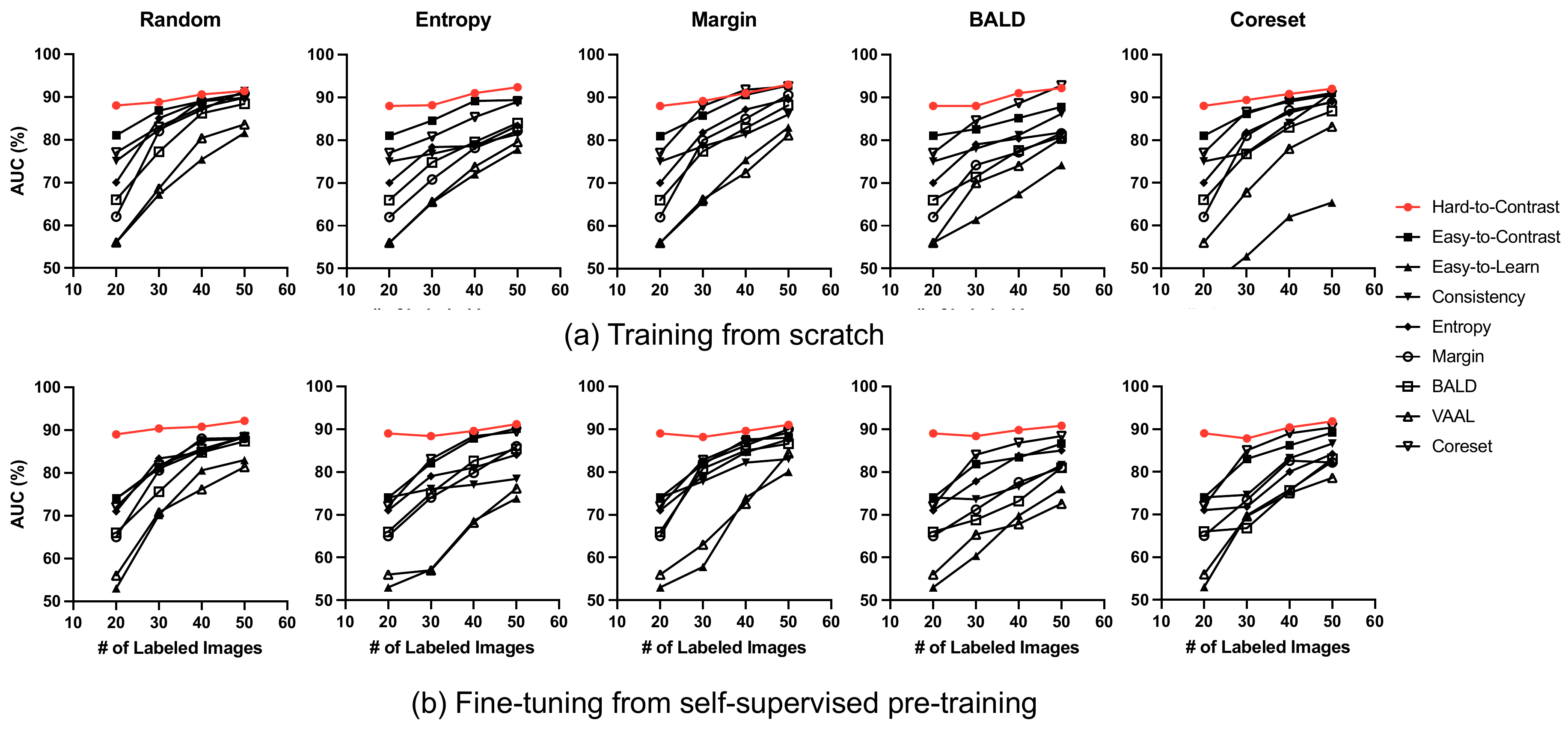}
    \caption{
        \textbf{Performance of each active learning querying strategies with different initial query strategies on BloodMNIST.}
        Hard-to-contrast initial query strategy (red lines) outperforms other initial query strategies in every cycle of active learning. With each active learning querying strategy, the performance of the initial cycle (20 labeled images) and the last cycle (50 labeled images) are strongly correlated. 
    }
    \label{fig:better_better_bloodmnist}
\end{figure}

\begin{figure}[!t]
    \includegraphics[width=\linewidth]{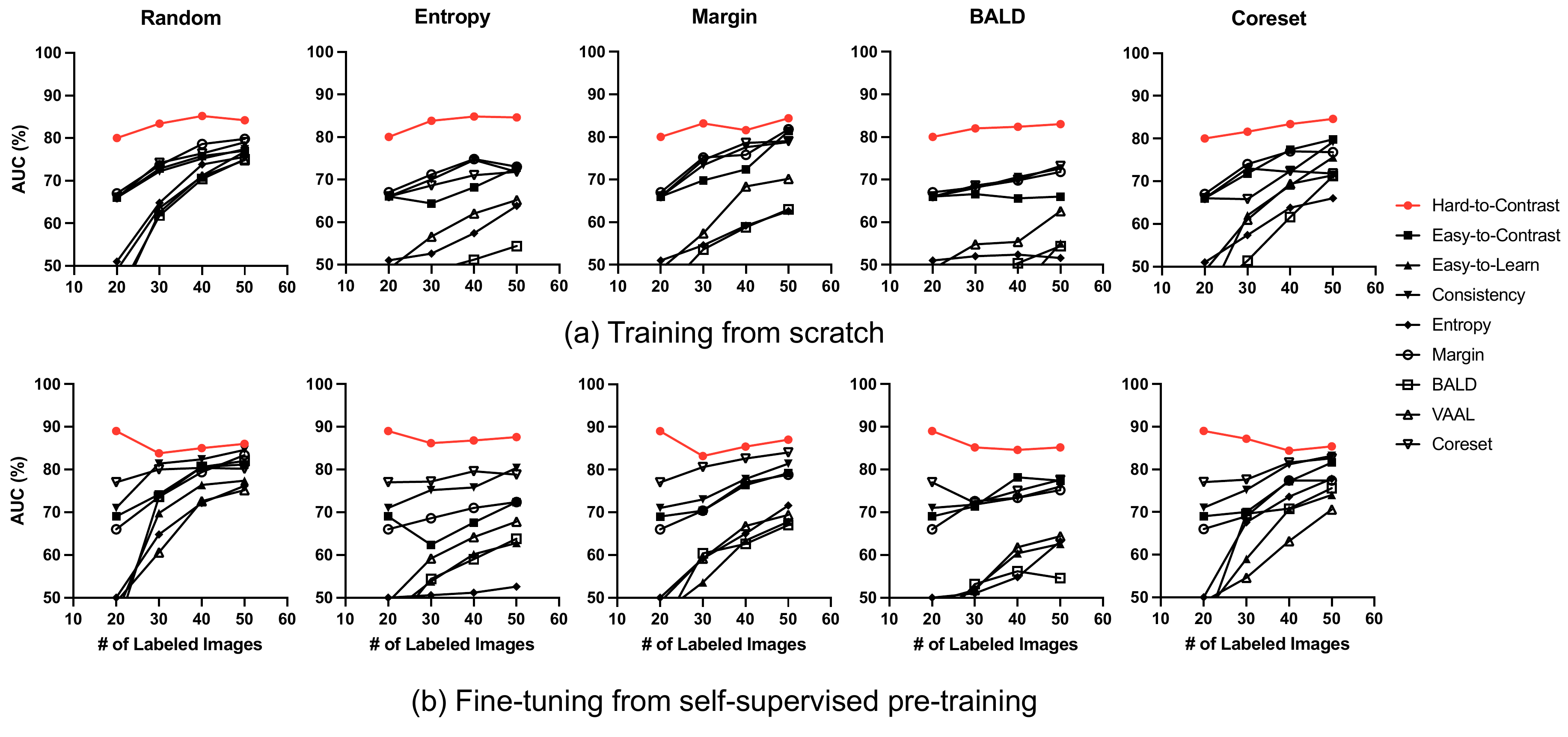}
    \caption{
        \textbf{Performance of each active learning querying strategies with different initial query strategies on PathMNIST.}
        Hard-to-contrast initial query strategy (red lines) outperforms other initial query strategies in every cycle of active learning. With each active learning querying strategy, the performance of the initial cycle (20 labeled images) and the last cycle (50 labeled images) are strongly correlated. 
    }
    \label{fig:better_better_pathmnist}
\end{figure}



\end{document}